\documentclass{cci}

\usepackage[utf8]{inputenc}

\usepackage[T1]{fontenc}
\usepackage{graphicx}
\usepackage{footmisc} 
\usepackage{titlesec}
\makeatletter
\let\AB@footnotetext\@footnotetext
\makeatother

\usepackage{authblk}

\title{AI4Math: A Native Spanish Benchmark for University-Level Mathematical Reasoning in Large Language Models}

\author[1,2,5]{Miguel Angel Peñaloza Perez}
\author[1,2,3]{Bruno Lopez Orozco}
\author[1,2,4]{Jesus Tadeo Cruz Soto}
\author[1,2]{Michelle Bruno Hernandez}
\author[1,2]{Miguel Angel Alvarado Gonzalez}
\author[1,2]{Sandra Malagon} 

\affil[1]{Carreras con Impacto}
\affil[2]{Aixo Lab}
\affil[3]{Facultad de Ciencias, UNAM, México.}
\affil[4]{Facultad de Matemáticas, Universidad Veracruzana, México.}
\affil[5]{Centro de Investigación Científica y de Educación Superior de Ensenada, Baja California, México.}

\date{}

\begin{document}

\maketitle

\footnotetext[1]{\textbf{Corresponding author:} \texttt{sandra.malagon@carrerasconimpacto.org}. 
\textbf{Other authors' emails:} \texttt{penaloza@cicese.mx}, 
\texttt{bruno.lo@ciencias.unam.mx}, 
\texttt{zS23014570@estudiantes.uv.mx}, 
\texttt{michelle.bruno@carrerasconimpacto.org}, 
\texttt{miguel.alvarado@carrerasconimpacto.org}.}

\thispagestyle{firstpage}
\begin{abstract}
Existing mathematical reasoning benchmarks are predominantly English-only or translation-based, which can introduce semantic drift and mask language-specific reasoning errors. To address this, we present AI4Math, a benchmark of 105 original university-level math problems natively authored in Spanish. The dataset spans seven advanced domains (Algebra, Calculus, Geometry, Probability, Number Theory, Combinatorics, and Logic), and each problem is accompanied by a step-by-step human solution. 
We evaluate six large language models — GPT-4o, GPT-4o mini, o3-mini, LLaMA 3.3 70B, DeepSeek-R1 685B, and DeepSeek-V3 685B — under four configurations: zero-shot and chain-of-thought, each in Spanish and English. The top models (o3-mini, DeepSeek-R1 685B, DeepSeek-V3 685B) achieve over 70\% accuracy, whereas LLaMA 3.3 70B and GPT-4o mini remain below 40\%. Most models show no significant performance drop between languages, with GPT-4o even performing better on Spanish problems in the zero-shot setting. Geometry, Combinatorics, and Probability questions remain persistently challenging for all models. These results highlight the need for native-language benchmarks and domain-specific evaluations to reveal reasoning failures not captured by standard metrics.
    
\end{abstract}


\fancypagestyle{plain}{
  \fancyhf{} 
  \renewcommand{\headrulewidth}{0pt}
}
\pagestyle{plain}
\newpage

\section{Introduction}

Benchmarking has become a cornerstone of evaluating large language models (LLMs), particularly in complex domains such as mathematical reasoning. Benchmarks such as MATH \citep{hendrycks2021math}, GSM8K \citep{cobbe2021}, and FrontierMath \citep{glazer2024} have been used to evaluate symbolic manipulation and multi-step problem solving, primarily in English. More recent benchmarks like MathVista \citep{lu2023mathvista} and Humanity’s Last Exam \citep{zheng2023humanits} further stress models with visual prompts or unsolved graduate-level questions. However, these evaluations remain entirely monolingual and reflect anglophone mathematical traditions.
\\
Benchmarks for mathematical reasoning in Spanish remain scarce. Existing efforts include translated versions of GSM8K and a recent study by \citep{parra2024geometry} evaluated LLM performance on a high-school geometry problem originally posed in Spanish, revealing significant conceptual and spatial reasoning errors. 
\\
This gap is particularly problematic in multilingual evaluation. Benchmarks like MMLU \citep{hendrycks2020mmlu}, Global MMLU \citep{singh2024globalmmlu}, and WorldBench \citep{moayeri2024worldbench} include Spanish among many languages but are built from English originals. Translation-based evaluation often introduces semantic drift, cultural mismatch, and unnatural phrasing. As \citep{plaza2024} show, translating MMLU into Spanish produces erratic results and semantic ambiguity. \citep{singh2024globalmmlu} similarly warn that translated benchmarks can unfairly penalize models not trained on English-centric curricula or logic structures. This is especially critical for mathematics, where meaning hinges on formal syntax and precise terminology.
\\
Designing rigorous, multilingual benchmarks also faces structural challenges. As \citep{kaddour2023llmchallenges}  observe, the high cost of development, annotation bottlenecks, and dependence on a small number of institutions pose barriers to transparent, inclusive evaluation. Most benchmarks are constructed by teams with institutional access to compute, funding, and alignment infrastructure. Addressing this imbalance requires collaborative methodologies that reduce cost and decentralize ownership. Equally important is the role of independent evaluation. As \citep{solaiman2024impact} argue, internal benchmarking lacks accountability, and a shift toward distributed, transparent evaluation frameworks is critical for trustworthy model assessment.
\\
In this paper, we introduce AI4Math, a university-level benchmark natively authored in Spanish for evaluating mathematical reasoning in LLMs. It includes 105 original problems spanning seven advanced domains—Algebra, Calculus, Geometry, Probability, Number Theory, Combinatorics, and Mathematical Logic—each with an exact answer and a full step-by-step human-written solution. Problems were collaboratively authored and reviewed by Latin American STEM students during a structured hackathon. The benchmark is designed to expose both language-driven disparities and domain-specific limitations across evaluation settings. AI4Math also exemplifies a community-led, peer-reviewed model for benchmark creation—providing a replicable, low-cost alternative to traditional development pipelines. As part of a broader movement toward native-language and diagnostic benchmarks, AI4Math complements efforts like Uhura \citep{uhura2024}, Te Reo Māori \citep{duncan2024}, and Command R+ \citep{cohere2024}, while contributing to a more equitable, multilingual evaluation ecosystem.

\section{Dataset and Benchmark Design}

\subsection{Dataset Construction}

To construct a high-quality mathematical reasoning benchmark originally authored in Spanish, we organized a two-day hackathon titled the AI4Math Challenge. This event brought together 6 undergraduate, master’s, and doctoral students in STEM disciplines across Latin America. Working in small teams, participants were tasked with designing original math problems across seven core domains, explicitly aimed at evaluating the reasoning capabilities of large language models (LLMs). The creation process was guided by detailed authoring protocols and followed a multi-stage peer review process to ensure quality, originality, and clarity.
\newline

\noindent{\textbf{Problem Domains and Guidelines:}} We targeted seven core areas of undergraduate-level mathematics: Algebra, Calculus, Geometry, Probability, Number Theory, Combinatorics, and Mathematical Logic. The challenge for participants was to formulate problems in these domains that were original (not copied from existing sources), clearly stated in Spanish, and solvable with well-defined correct answers. To standardize submissions, we provided the following guidelines for each problem:

\begin{itemize}
    \item The problem statement should be concise (under 120 words) and use standard mathematical notation (including LaTeX for clarity when needed). Unnecessary symbols or convoluted language were discouraged.
    \item Each problem must have a unique correct answer (e.g. a specific number, expression, or choice) so that automated checking is straightforward. Ambiguous questions or those with multiple valid answers were not allowed.
    \item The solution to the problem should be verifiable without excessive computation. In other words, problems requiring brute-force or extremely lengthy calculations were avoided, to focus on reasoning over raw computation.
    \item The difficulty should be at roughly university undergraduate level or slightly lower. Extremely specialized or research-level questions were out of scope; the goal was to create challenging yet solvable problems that an advanced student might encounter.
   \item Along with the problem statement, participants had to provide a detailed step-by-step solution culminating in the final answer. This solution served as the ground-truth for evaluation and for verifying model outputs. The solution steps were expected to be logically sound and to clearly explain the reasoning. 
\end{itemize}

To encourage breadth, each participant could submit at most three problems per area. This helped prevent any single contributor or area from dominating the dataset, and encouraged participants to focus on quality over quantity.

\noindent\textbf{Collaborative Review Process:} To ensure the validity and clarity of each problem, we implemented a multi-step peer review and validation pipeline. In the first round, authors anonymously exchanged problems for cross-review, evaluating them for clarity, correctness, and adherence to the guidelines. Submissions flagged as ambiguous, unsolvable, or inconsistent were revised by the original author. A second round of review by a different team ensured unbiased quality control. Only problems that passed both reviews were considered for inclusion in the benchmark. This community-driven vetting process ensured high inter-annotator agreement and eliminated questions with ambiguous wording or multiple possible answers.

As a final validation step, we performed a live diagnostic evaluation during the hackathon using GPT-4o. Each curated problem was presented to the model in real-time, and participants were awarded a point if the model failed to solve it correctly. This human–AI comparison served both as a competitive incentive and a stress test to identify overly simple or flawed problems. Problems trivially solved by the model were flagged for potential revision or removal. The most difficult problems (those consistently missed by GPT-4o) were recognized with awards, encouraging contributions that challenged state-of-the-art reasoning capabilities.

\noindent\textbf{Dataset Composition:} The final dataset comprises 105 problems, balanced across the seven mathematical domains (see Table 1). Each problem entry includes: (1) the original Spanish problem statement, (2) the exact final answer, and (3) a detailed, human-authored step-by-step solution. The inclusion of full solutions is a distinguishing feature of AI4Math, enabling both final-answer evaluation and in-depth analysis of model reasoning processes. Problems were intentionally designed to elicit reasoning rather than brute-force calculation, and all were vetted for correctness, clarity, and solvability. The structure of the dataset makes it suitable for evaluating both direct answer generation and chain-of-thought performance.

\begin{table}[h!]
\centering
\begin{tabular}{|l|c|c|}
\hline
\textbf{Domain} & \textbf{Number of Problems} & \textbf{Percentage} \\
\hline
Algebra            & 15 & 14.29\% \\
Calculus           & 15 & 14.29\% \\
Combinatorics      & 14 & 13.33\% \\
Geometry           & 15 & 14.29\% \\
Mathematical Logic & 15 & 14.29\% \\
Number Theory      & 16 & 15.24\% \\
Probability        & 15 & 14.29\% \\
\hline
\textbf{Total}     & \textbf{105} & \textbf{100.00\%} \\
\hline
\end{tabular}
\caption{Distribution of problems by mathematical domain. This table presents the distribution of the 105 problems in the AI4Math dataset across seven mathematical domains. Each problem includes a unique, exact answer and a detailed step-by-step solution. Percentages indicate the relative representation of each domain within the dataset.}
\label{tab:distribution}
\end{table}

\subsubsection*{Example}

\subsubsection*{Domain: Probability}

\begin{itemize}
    \item Spanish Version: Tenemos tres dados, uno es un icosaedro, otro es un cubo, y el último es un tetraedro. ¿Cuál es la probabilidad de que al tirar los tres dados la suma no sea 2? 

    \item English Version: We have three dice, one is an icosahedron, one is a cube, and the last one is a tetrahedron. What is the probability that when rolling the three dice the sum is not 2?

    \item Answer: 100\%
\end{itemize}

\subsubsection*{Domain: Algebra}

\begin{itemize}
    \item Spanish Version: Una "palabra" consiste de concatenaciones finitas de simbolos en un alfabeto. Si el alfabeto es el conjunto $\{a,b,c,d,e\}$ ¿Cuantas palabras con exactamente dos a's hay?

    \item English Version: A "word" consists of finite concatenations of symbols in an alphabet. If the alphabet is the set $\{a,b,c,d,e\}$ how many words with exactly two a's are there?

    \item Answer: infinite
\end{itemize}

\subsubsection*{Domain: Geometry}

\begin{itemize}
    \item Spanish Version: Se tiene medio círculo cuyo diámetro es $10$, dicho semicírculo tiene dos cuadrados con lados $x$ y $y$, alineados lado a lado y con el diámetro, uno de los vértices de cada cuadrado toca el arco del círculo. Calcula el área de los dos cuadrados.

    \item English Version: We have a half circle whose diameter is $10$, this semicircle has two inscribed squares with sides $x$ and $y$, aligned side by side and with the diameter, one of the vertices of each square touches the arc of the circle and one of the vertices of each square is the center of the semicircle. Calculate the sum of the areas of both squares.

    \item Answer: $40$
\end{itemize}

\subsubsection*{Domain: Calculus}

\begin{itemize}
    \item Spanish Version:  Una esfera sólida pesa p en kg, ¿Cuál es el peso del mayor cilindro circular recto que puede cortarse de la esfera?

    \item English Version: A solid sphere weighs $p$ in kg, what is the weight of the largest straight circular cylinder that can be cut from the sphere?

    \item Answer: $\frac{3p}{4\sqrt{2}}$ Kg
\end{itemize}

\textbf{Dataset Availability:} To mitigate the risk of data contamination \citep{dong2024},  \citep{xu2024contamination} in future training runs, the AI4Math  dataset is initially available upon request. This restricted release strategy ensures controlled access for research purposes while preserving the benchmark’s long-term diagnostic value. Once a model has been trained on a benchmark, its utility as a test set may be compromised. As such, we discourage wide-scale redistribution and instead support researchers through a documented access process. The dataset includes metadata for each item—domain labels, difficulty ratings, and verification history—to facilitate reproducibility and secondary analysis. Full release under an academic-use license is planned pending further stability testing and downstream evaluations.

\section{Evaluation Setup
}

We evaluated six state-of-the-art LLMs on the AI4Math benchmark under uniform experimental conditions. These models represent a mix of closed and open systems, as well as general-purpose and reasoning-specialized architectures: GPT-4o, GPT-4o mini, o3 mini, LLaMA 3.3 70B, DeepSeek-R1 685B and DeepSeek-V3 685B.

\noindent\textbf{Evaluation Conditions:} Language and Reasoning Setup: To assess the reasoning capabilities of large language models on AI4Math, we designed four evaluation conditions that vary along two axes: input language (Spanish vs. English) and direct answer vs. chain-of-thought.\\
These conditions allow us to isolate the effects of linguistic competence and step-by-step reasoning guidance.

\begin{itemize}
    \item \textbf{Zero-Shot in Spanish (ZS-ES):} The original problem, written in Spanish, is presented to the model without additional instructions or examples. This condition reflects a model’s unassisted ability to interpret and solve problems in a non-English language.
    \item \textbf{Zero-Shot in English (ZS-EN):} The problem is translated to English through automatic translation using the DeepL API (DeepL v2 AP, 2025) followed by a  thorough bilingual review by the authors. This approach isolates the effect of language on performance, as the mathematical content remains constant.
    \item \textbf{Zero-Shot Chain-of-Thought in Spanish (ZS-CoT-ES):} The model is prompted to reason step-by-step in Spanish using a prefix such as “Explica tu respuesta paso a paso:”. This encourages explicit intermediate reasoning in the model’s output.
    \item \textbf{Zero Shot Chain-of-Thought in English (ZS-CoT-EN):} The English-translated problem is preceded by “Explain your answer step by step:” to induce structured reasoning in English.
    
\end{itemize}

\noindent\textbf{Evaluation:} All model outputs were manually reviewed by the authors. Final correctness was determined based solely on the match between the model’s final answer and the ground-truth solution.

\noindent\textbf{Token Budget and Other Settings:} All model responses were generated using a fixed decoding configuration across all conditions to ensure comparability. We used a temperature of 0.7 and top-p of 0.95, selected to encourage coherent reasoning while allowing moderate variation in output style.

A maximum output length of 2,500 tokens was allowed to ensure that models could generate complete step-by-step solutions under both Chain-of-Thought and zero-shot settings. Although no explicit constraint was imposed, human-written solutions typically remained under 800 tokens.

The AI4Math dataset creation process involved the development of problems by human experts through a rigorous, multi-stage process involving collaborative writing and peer review. The final set of 105 Spanish problems was then presented to various LLMs under different evaluation strategies. Each model’s solutions — including potential step-by-step reasoning — were verified through human checks, and the results were subsequently aggregated for analysis.  

\noindent\textbf{The statistical analysis} of the performance evaluations was conducted in POSIT Cloud 2022. To verify significant differences in the performace of LLMs across different configurations and languages, a McNemar Test and $\chi^2$  test was applied with a significance level of 0.05. The statistical tests were performed using the rstatix library and R base, and Holm´s method was used for post hoc p-value adjustment.

\section{Results}

This section reports the performance of six large language models (LLMs) on the AI4Math benchmark across four evaluation conditions: Zero-Shot (ZS) and Zero-Shot Chain-of-Thought (ZS-CoT), each in English and Spanish. Following the taxonomy proposed by \citep{schulhoff2024promptreport}, ZS settings present models with only the problem statement, while ZS-CoT includes a minimal reasoning prompt (e.g., “Explain your answer step by step:” or “Explica tu respuesta paso a paso:”) to encourage intermediate steps without providing exemplars.

Model performance is measured by success rate, defined as the percentage of problems where the final answer exactly matches the reference solution. All results are reported per model and configuration, based on a single run per condition, without ensembling or retries.

Across the 105 benchmark problems, o3-mini and DeepSeek-R1 685B were the top performers, consistently achieving over 70\% accuracy. In contrast, GPT-4o mini and LLaMA 3.3 70B obtained the lowest scores, remaining below 40\% in most configurations. These differences are detailed in Fig.  \ref{fig:scatterplot}, Tab. \ref{table:McNemarZS} and Tab \ref{table:McNemarCoT}.

To assess intra-model variation across configurations and languages, we applied McNemar’s test at a significance level of 0.05. For five of the six models, performance differences between languages and setting modes were not statistically significant. The sole exception was GPT-4o, which performed significantly better in Spanish than in English under the ZS setting (p = 0.008) (Tab. \ref{table:McNemarZS}).

Inter-model differences were more pronounced. Chi-square and post hoc analyses revealed that o3-mini, DeepSeek-R1 685B, and DeepSeek-V3 685B significantly outperformed the remaining models across conditions. However, performance differences among these three top models were not statistically significant (p $>$ 0.05), indicating comparable capabilities under the current evaluation. The post hoc test statistics are provided in the Appendix (Tables 7-10).

\subsection{\textbf{Overall Performance by Language and Evaluation Settings}}

\begin{figure}[H]
    \begin{center}
        \includegraphics[width=0.9\textwidth]{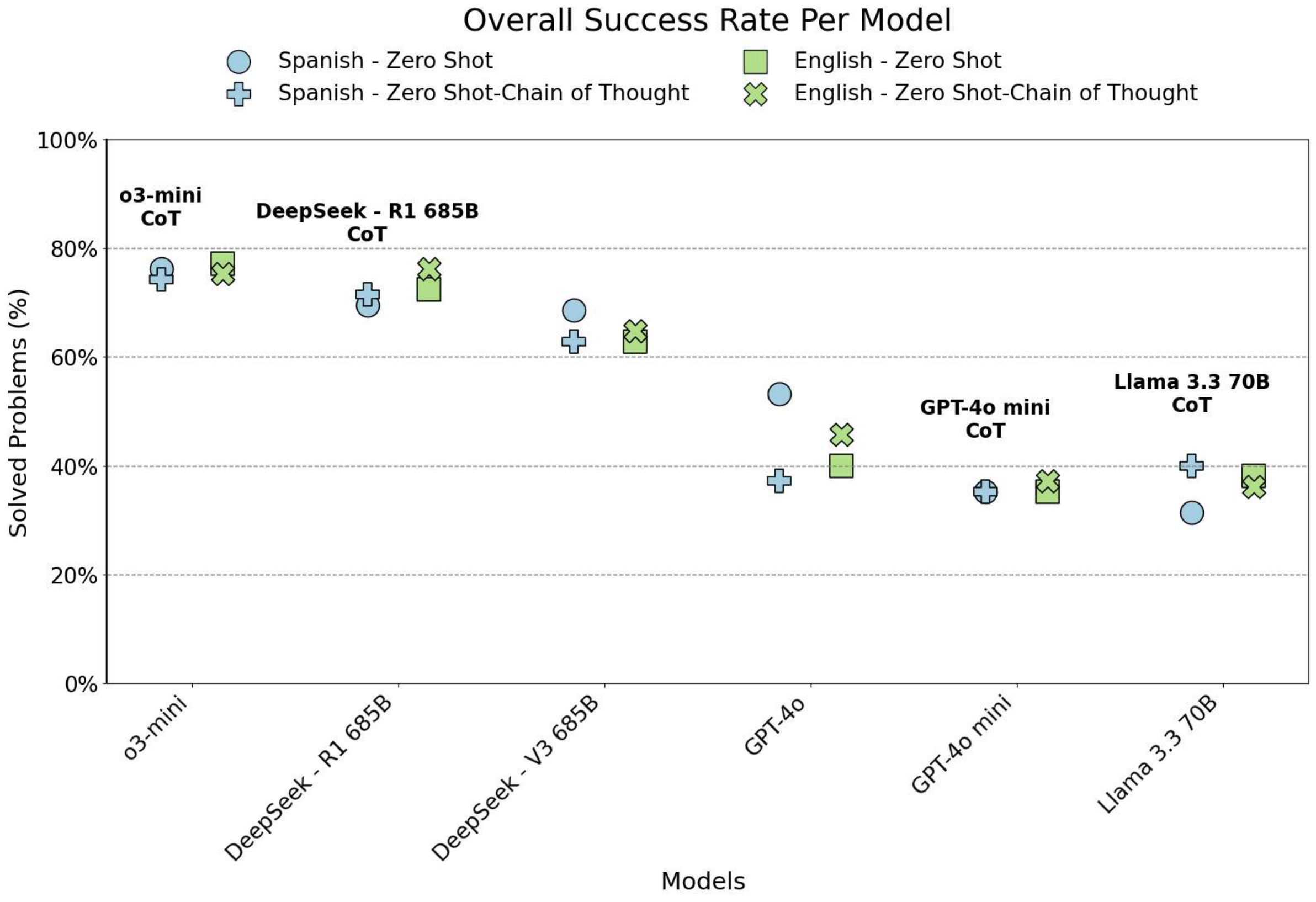}
    \end{center}
    \caption{Overall success rate per model. The figure presents the model names on the X-axis (o3-mini, GPT-4o, GPT-4o mini, DeepSeek-R1 685B, DeepSeek-V3 685B, LLaMA 3.3 70B) and the percentage of successfully solved problems on the Y-axis. For each model, the success rate is categorized by evaluation settings (Zero Shot and Zero Shot - Chain of Thought) and by language (Spanish and English). The comparison highlights performance differences across models and evaluation configurations, illustrating how each model's success rate varies between languages and reasoning approaches. }\label{fig:scatterplot}
\end{figure}

Notably, o3-mini solved 76\% of problems in Spanish-Zero Shot, while DeepSeek-R1 685B achieved 76.2\% in English-Zero Shot-Chain of Thought. In contrast, LLaMA 3.3 70B and GPT-4o mini showed consistently lower performance across all configurations, typically under 40\%. These results underscore both architectural differences and the importance of reasoning guidance and language support (Fig. \ref{fig:scatterplot}).

\begin{table}[H]
    \centering
    \begin{tabular}{|l|c|c|c|c|}
    \hline
    \textbf{LLM} & \textbf{English\_ZS} & \textbf{Spanish\_ZS} & \textbf{Statistic (df=1)} & \textbf{$p$-value} \\
    \hline
    GPT-4o & 40.00\% & 53.33\% & 7.0417 & $0.00796$ \textbf{**} \\
    \hline
    GPT-4o mini & 35.24\% & 35.24\% & 0 & $1$  \\
    \hline
    o3 mini & 77.14\% & 76.19\% & 0 & $1$  \\
    \hline
    LLaMA 3.3 70B & 38.10\% & 31.43\% & 1.33 & $0.2482$  \\
    \hline
    DeepSeek-R1 685B & 72.38\% & 69.52\% & 0.8 & $0.3711$  \\
    \hline
    DeepSeek-V3 685B & 62.86\% & 68.57\% & 1.5625 & $0.2113$  \\
    \hline
    \end{tabular}
    \caption{Overall success rate per model. Summarizes the overall success rate of each model in both Spanish and English under Zero Shot (ZS) configurations. The o3-mini and DeepSeek-R1 685B models were the top performers across both languages. Furthermore, the table presents the results of McNemar tests for paired samples across both languages under the Zero-Shot configuration, computed with 1 degree of freedom and a 95\% confidence level. Asterisks denote significance levels (* $p<0.05$, ** $p<0.01$, *** $p<0.001$).}
    \label{table:McNemarZS}
\end{table}

\begin{table}[H]
\centering
\begin{tabular}{|l|c|c|c|c|}
\hline
\textbf{LLM} & \textbf{English\_CoT} & \textbf{Spanish\_CoT} & \textbf{Statistic (df=1)} & \textbf{$p$-value} \\
\hline
GPT-4o & 45.71\% & 37.14\% & 2.3704 & $0.1237$  \\
\hline
GPT-4o mini & 37.14\% & 35.24\% & 0.04 & $0.83$  \\
\hline
o3 mini & 75.24\% & 74.29\% & 0 & $1$  \\
\hline
LLaMA 3.3 70B & 36.19\% & 40.00\% & 0.34615 & $0.5563$  \\
\hline
DeepSeek-R1 685B & 76.19\% & 71.43\% & 1.238 & $0.2673$  \\
\hline
DeepSeek-V3 685B & 64.76\% & 62.86\% & 0.0625 & $0.8026$  \\
\hline
\end{tabular}
\caption{Comparison of model performance between {English Zero Shot Chain of Thought} and {Spanish Zero Shot Chain of Thought}. The table shows success rates and McNemar test results for language-level differences under the Zero-Shot Chain-of-Thought configuration. Asterisks denote significance levels (* $p<0.05$, ** $p<0.01$, *** $p<0.001$).}
\label{table:McNemarCoT}
\end{table}

\begin{table}[H]
    \centering
    \begin{tabular}{|l|c|c|c|}
    \hline
    \textbf{Setting} & \textbf{$\chi^2$ (df=5)} & \textbf{$p$-value} \\
    \hline
    English\_ZS & $74.1$ & $1.4 \times 10^{-14}$  \textbf{***}  \\
    \hline
    Spanish\_ZS & $76.2$ & $ 5.29 \times 10^{-15}$  \textbf{***}  \\
    \hline
    English\_ZS\_CoT & $72.7$ & $2.74 \times 10^{-14}$  \textbf{***}  \\
    \hline
    Spanish\_ZS\_CoT & $68.6$ & $ 2.05 \times 10^{-13}$  \textbf{***}  \\
    \hline
    \end{tabular}
    \caption{Chi-squared tests for global comparison of model performance across languages and prompting configurations. The table reports the test statistic, $p$-value, and significance level for each condition. Asterisks denote significance levels (* $p<0.05$, ** $p<0.01$, *** $p<0.001$).}\label{table:ChiSquaredGlobal}
\end{table}

\subsection{Performance by Domain: English
}

The following tables present the model performance by domain under the AI4Math benchmark for English configurations.

\begin{figure}[H]
    \begin{center}
        \includegraphics[width=0.9\textwidth]{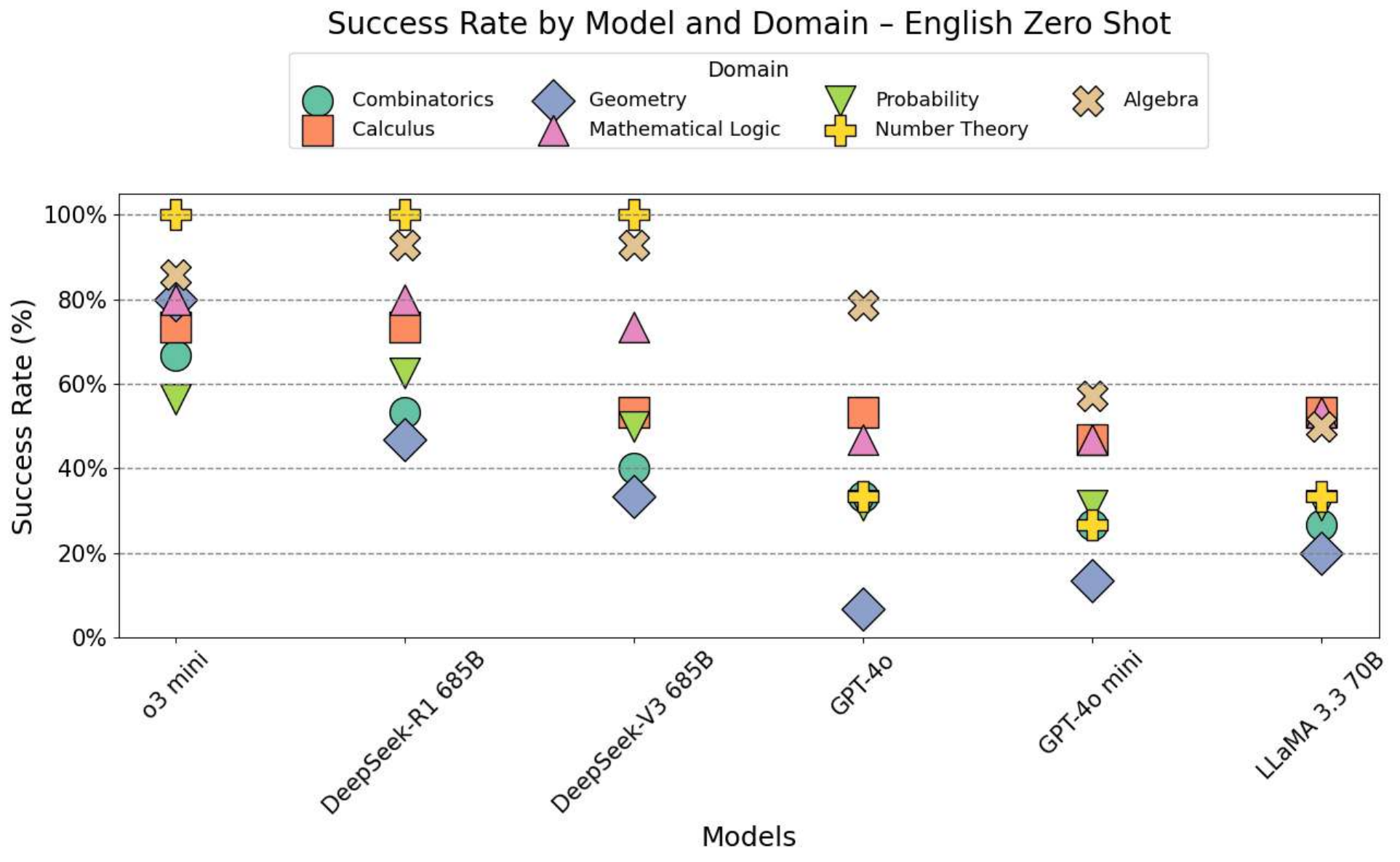}
    \end{center}
    \caption{The figure presents the percentage of correctly solved problems by model and domain under the English Zero-Shot configuration. Each marker represents the performance of a specific model across seven mathematical domains: Combinatorics, Calculus, Geometry, Mathematical Logic, Probability, Number Theory, and Algebra. Tab.\ref{Table:domainChiSqauredEnglishZS} and Tab. \ref{Tab:domainChiSqauredEnglishZSModels} report the statistical differences observed across models and mathematical domains.}\label{fig:Zs_English}
\end{figure}

Additionally, inter-model performance differences were assessed across mathematical domains. According to Chi-Squared tests with a 0.05 significance level (Tab. \ref{Table:domainChiSqauredEnglishZS}), significant differences were observed in Number Theory ($\chi^2$=47.6, df=5, p$<$0.05), Geometry ($\chi^2$=24.6, df=5, p$<$0.05), and Algebra ($\chi^2$=13.2, df=5, p$<$0.05). Detailed post hoc results are provided in the appendices (Appendix Tab. \ref{Tab:PosthoctestresultsforGeometryundertheEnglish_ZSconfiguration.}, \ref{Tab:PosthoctestresultsforNumberTheoryundertheEnglish_ZSconfiguration.}, \ref{Tab:PosthoctestresultsforAlgebraundertheEnglish_ZSconfiguration.})

\begin{table}[H]
\centering
\begin{tabular}{|l|c|c|l|}
\hline
\textbf{Area} & \textbf{$\chi^2$ (df=5)} & \textbf{$p$-value} \\
\hline
Combinatorics     & 7.94  & 0.16            \\
\hline
Calculus          & 4.08  & 0.537           \\
\hline
Geometry          & 24.6  & 0.000166       \textbf{***}  \\
\hline
Mathematical Logic& 8.47  & 0.132           \\
\hline
Probability       & 6.60  & 0.252           \\
\hline
Number Theory     & 47.6  & $4.52 \times 10^{-9}$  \textbf{***}  \\
\hline
Algebra           & 13.2  & 0.0222         \textbf{*}  \\
\hline
\end{tabular}
\caption{Chi-squared test results for model comparison per mathematical domain under the English Zero Shot setting. Asterisks denote significance levels (* $p<0.05$, ** $p<0.01$, *** $p<0.001$).} \label{Table:domainChiSqauredEnglishZS}
\end{table}

Chi-Squared tests were applied to identify significant intra-model differences across domains at a 0.05 significance level. According to the test results, significant intra-model differences were identified for GPT-4o ($\chi^2$=18.1, df=6, p=0.00604), DeepSeek-R1 685B ($\chi^2$=17.6, df=6, p=0.00741), and DeepSeek-V3 685B ($\chi^2$=25.6, df=6, p=0.00026). (Tab. \ref{Tab:domainChiSqauredEnglishZSModels}). Detailed post hoc test results are included in the appendices (Tab. \ref{Tab:PosthoctestresultsforGPT-4obydomainundertheEnglish_ZSconfiguration.}, \ref{Tab:PosthoctestresultsforDeepSeek-R1bydomainundertheEnglish_ZSconfiguration.}, \ref{Tab:PosthoctestresultsforDeepSeek-V3bydomainundertheEnglish_ZSconfiguration.}).

\begin{table}[H]
\centering
\caption{Chi-squared test results comparing success rates across domains for each model under the English Zero Shot setting. Asterisks denote significance levels (* $p<0.05$, ** $p<0.01$, *** $p<0.001$).}\label{Tab:domainChiSqauredEnglishZSModels}
\begin{tabular}{|l|c|c|l|}
\hline
\textbf{Model} & \textbf{$\chi^2$ (df=6)} & \textbf{$p$-value} \\
\hline
GPT-4o            & 18.1  & 0.00604      \textbf{***} \\
\hline
GPT-4o mini       & 8.89  & 0.18          \\
\hline
o3 mini           & 10.2  & 0.117         \\
\hline
LLaMA 3.3 70B     & 7.17  & 0.305         \\
\hline
DeepSeek-R1 685       & 17.6  & 0.00741      \textbf{**}  \\
\hline
DeepSeek-V3 685       & 25.6  & 0.00026      \textbf{***} \\
\hline
\end{tabular}
\end{table}

For the Fig. \ref{fig:Zs_English}, which evaluates model performance in the Chain-of-Thought configuration in English, significant intra-model differences were identified for DeepSeek-R1 685 ($\chi^2$=12.8, df=6, p=0.047) and DeepSeek-V3 685 ($\chi^2$=13.09, df=6, p=0.0311). (Tab.\ref{tab:domainChiSqauredEnglishCoTModels}). Moreover, significant inter-model differences were found in Number Theory ($\chi^2$=35, df=5, p$<$0.05), Geometry ($\chi^2$=15.7, df=5, p<0.05), and Mathematical Logic ($\chi^2$=12, df=5, p=0.035), as verified through Chi-Squared tests at a 95\% confidence level Tab. \ref{Table:domainChiSqauredEnglishCoT}. Post hoc analyses for these comparisons are included in the appendices (Appendix Tab. \ref{Tab:PosthoctestresultsforGeometryundertheSpanish_ZS_CoTconfiguration.}, \ref{tab:posthoc_mathlogic_english_zs_cot},   \ref{tab:posthoc_number_theory_english_zs_cot}, \ref{Tab:PosthoctestresultsforDeepSeek-R1bydomainundertheEnglish_ZS_CoTconfiguration.}, \ref{Tab:PosthoctestresultsforDeepSeek-V3bydomainundertheEnglish_ZS_CoTconfiguration.}).

\begin{figure}[H]
    \begin{center}
        \includegraphics[width=0.9\textwidth]{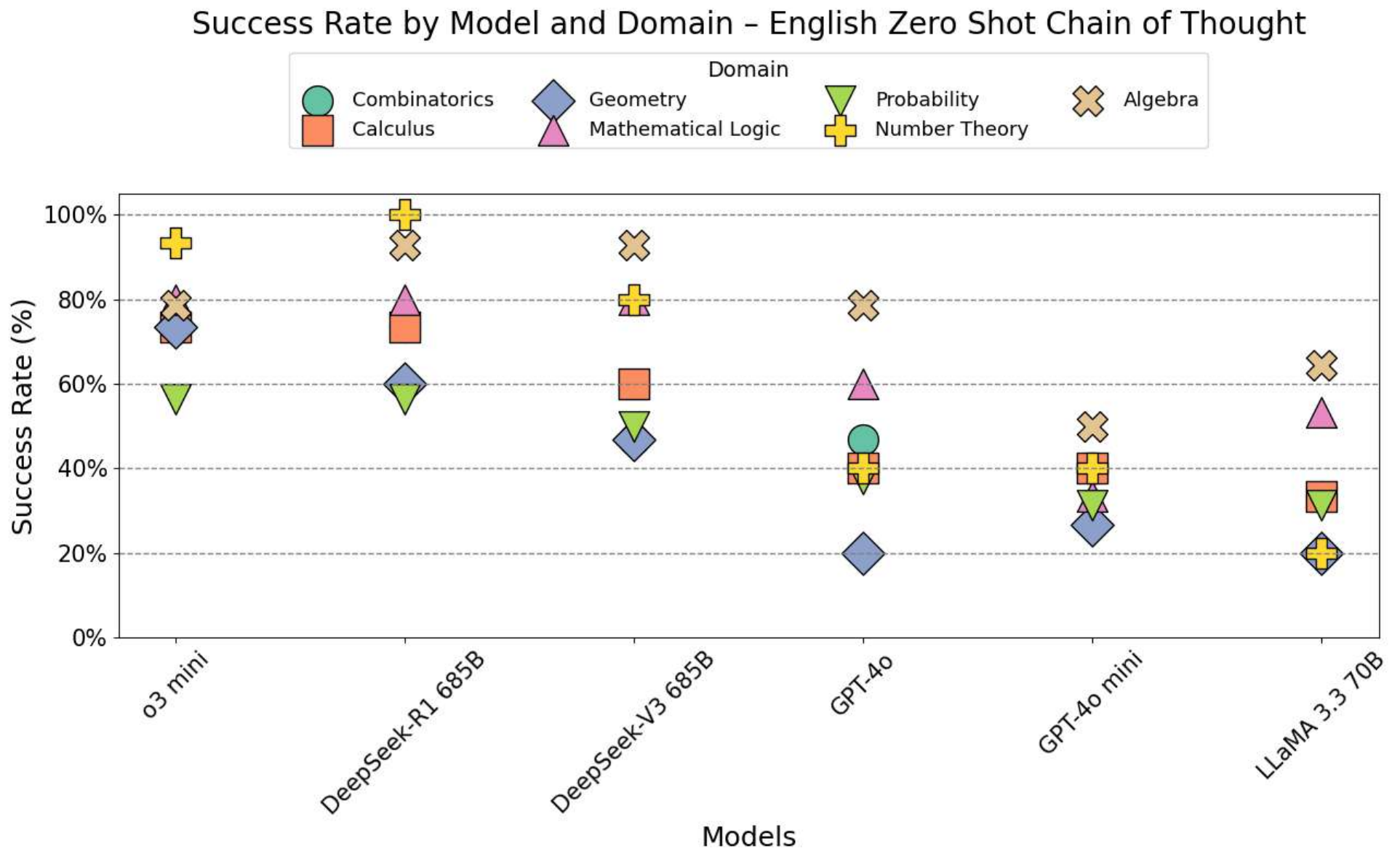}
    \end{center}
    \caption{Presents the percentage of correctly solved problems by model and domain under the English Zero-Shot Chain of Thought configuration. Each marker represents the performance of a specific model across seven mathematical domains: Combinatorics, Calculus, Geometry, Mathematical Logic, Probability, Number Theory, and Algebra.  Tab.\ref{Table:domainChiSqauredEnglishCoT} and Tab. \ref{tab:domainChiSqauredEnglishCoTModels} report the statistical differences observed across models and mathematical domains.}\label{fig:Zs_English}
\end{figure}

Overall, Algebra and Number Theory were the best-performing domains across all models and configurations. In contrast, Geometry, Combinatorics, and Probability had lower success rates and greater variance between models. Performance generally improved under Zero Shot-Chain of Thought prompting, though gains varied by domain and model architecture.

\begin{table}[H]
\centering
\caption{Chi-squared test results for model comparison per mathematical domain under the English Zero Shot Chain of Thought setting. Asterisks denote significance levels (* $p<0.05$, ** $p<0.01$, *** $p<0.001$).}\label{Table:domainChiSqauredEnglishCoT}
\begin{tabular}{|l|c|c|l|}
\hline
\textbf{Area} & \textbf{$\chi^2$ (df=5)} & \textbf{$p$-value}  \\
\hline
Combinatorics     & 8.77   & 0.118           \\
\hline
Calculus          & 9.64   & 0.086           \\
\hline
Geometry          & 15.7   & 0.00792        \textbf{**} \\
\hline
Mathematical Logic& 12.0   & 0.0344         \textbf{*}  \\
\hline
Probability       & 4.57   & 0.47            \\
\hline
Number Theory     & 35.0   & $1.52 \times 10^{-6}$  \textbf{***}  \\
\hline
Algebra           & 10.8   & 0.0563          \\
\hline
\end{tabular}
\end{table}

\begin{table}[H]
\centering
\caption{Chi-squared test results comparing performance across mathematical domains for each model under the English Zero Shot Chain of Thought setting. Asterisks denote significance levels (* $p<0.05$, ** $p<0.01$, *** $p<0.001$).}\label{tab:domainChiSqauredEnglishCoTModels}
\begin{tabular}{|l|c|c|l|}
\hline
\textbf{Model} & \textbf{$\chi^2$ (df=6)} & \textbf{$p$-value} \\
\hline
GPT-4o            & 12.2  & 0.0586         \\
\hline
GPT-4o mini       & 2.18  & 0.902          \\
\hline
o3 mini           & 6.09  & 0.414          \\
\hline
LLaMA 3.3 70B     & 10.4  & 0.11           \\
\hline
DeepSeek-R1 685B      & 12.8  & 0.047         \textbf{*}  \\
\hline
DeepSeek-V3 685B       & 13.09 & 0.0311        \textbf{*}  \\
\hline
\end{tabular}
\end{table}

\subsection{Performance by Domain: Spanish
}

The following section presents the performance of the evaluated LLMs by domain within the AI4Math benchmark under the Spanish configuration. 

\begin{figure}[H]
    \begin{center}
        \includegraphics[width=0.9\textwidth]{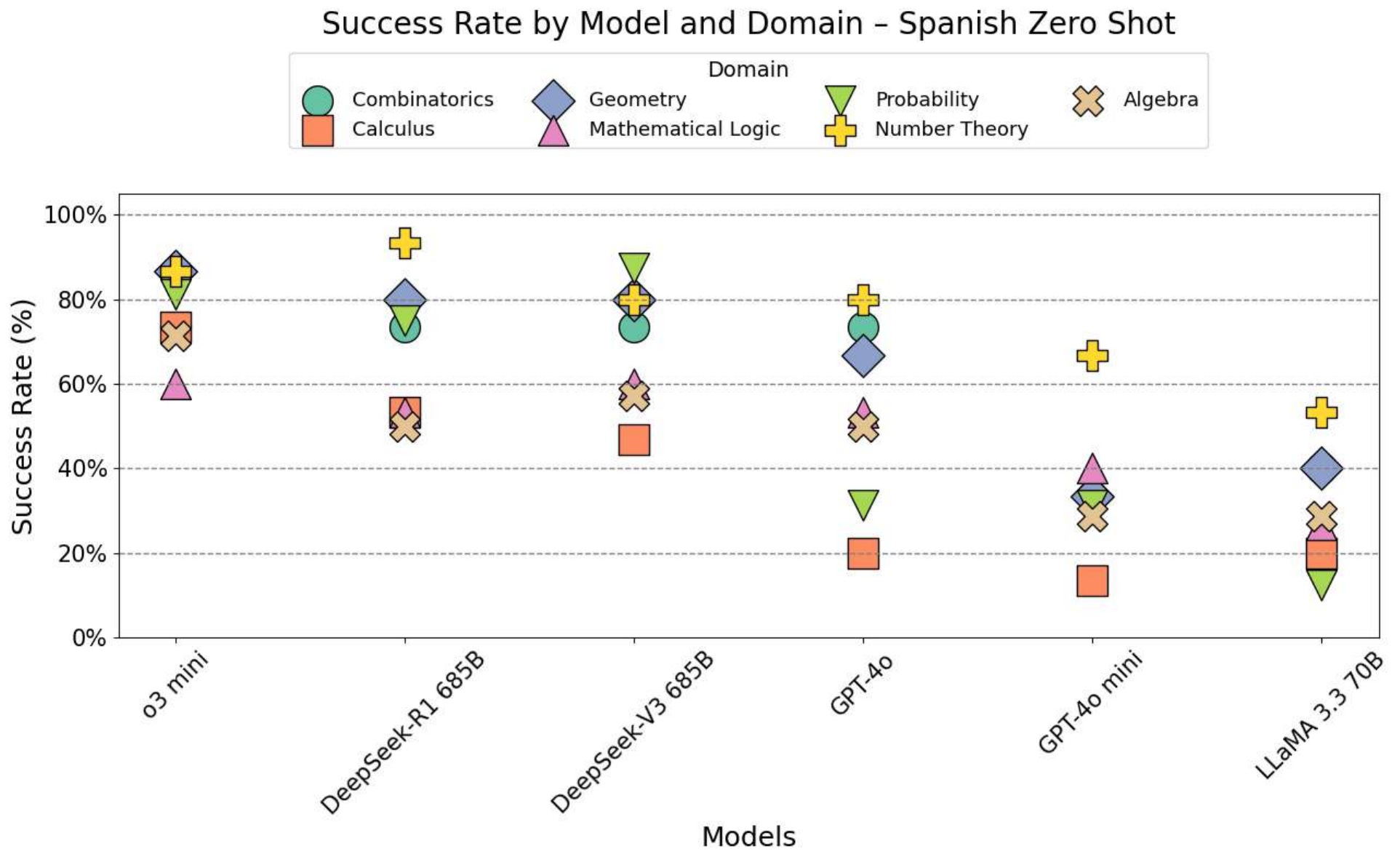}
    \end{center}
    \caption{presents the success rate of large language models across different mathematical domains under the Spanish Zero-Shot configuration. Each marker corresponds to a domain Combinatorics, Calculus, Geometry, Mathematical Logic, Probability, Number Theory, Algebra and shows the percentage of correctly solved problems by each model. Tab \ref{Tab:domainChiSqauredSpanishZS} and Tab. \ref{fig:domainChiSqauredSpanishZSModels} report the statistical differences observed across models and mathematical domains.}\label{fig:domainChiSqauredSpanishZS}
\end{figure}

Additionally, inter-model performance differences were identified in Combinatorics, Calculus, Geometry, and Probability, with a confidence level of 95\% and p-values below 0.05 (Tab. \ref{Tab:domainChiSqauredSpanishZS}). Detailed post hoc test results are provided in the appendices (Appendix Tab. \ref{Tab:PosthoctestresultsforCombonatoricsundertheSpanish_ZSconfiguration.}, \ref{Tab:PosthoctestresultsforCalculusundertheSpanish_ZSconfiguration.}, \ref{tab:posthoc_geometry_spanish_zs}, \ref{tab:posthoc_probability_spanish_zs}) .

\begin{table}[H]
\centering
\caption{Chi-squared test results for model comparison per mathematical domain under the Spanish Zero Shot setting. Asterisks denote significance levels (* $p<0.05$, ** $p<0.01$, *** $p<0.001$).}\label{Tab:domainChiSqauredSpanishZS}
\begin{tabular}{|l|c|c|l|}
\hline
\textbf{Area} & \textbf{$\chi^2$ (df=5)} & \textbf{$p$-value} \\
\hline
Combinatorics     & 11.5   & 0.0431         \textbf{*}  \\
\hline
Calculus          & 18.0   & 0.00299        \textbf{**}  \\
\hline
Geometry          & 16.7   & 0.00515        \textbf{**}  \\
\hline
Mathematical Logic& 5.16   & 0.397           \\
\hline
Probability       & 32.5   & $4.73 \times 10^{-6}$  \textbf{***}  \\
\hline
Number Theory     & 8.76   & 0.119           \\
\hline
Algebra           & 7.83   & 0.166           \\
\hline
\end{tabular}
\end{table}

Additionally, Chi-Squared tests were applied to assess intra-model and inter-model differences with a significance level of 0.05 across both Zero-Shot and Chain-of-Thought configurations in Spanish.
For the Zero-Shot configuration (Tab. \ref{fig:domainChiSqauredSpanishZSModels}, \ref{Tab:PosthoctestresultsforGPT-4obydomainundertheSpanish_ZSconfiguration.}), significant intra-model differences were observed across the seven mathematical domains for the GPT-4o model ($\chi^2$=17.7, df=6, p=0.00714). 

\begin{table}[H]
\centering
\caption{Chi-squared test results comparing performance across mathematical domains for each model under the Spanish Zero Shot setting. Asterisks denote significance levels (* $p<0.05$, ** $p<0.01$, *** $p<0.001$).}\label{fig:domainChiSqauredSpanishZSModels}
\begin{tabular}{|l|c|c|l|}
\hline
\textbf{Model} & \textbf{$\chi^2$ (df=6)} & \textbf{$p$-value} \\
\hline
GPT-4o            & 17.7  & 0.00714      \textbf{**}  \\
\hline
GPT-4o mini       & 10.2  & 0.115         \\
\hline
o3 mini           & 4.52  & 0.607         \\
\hline
LLaMA 3.3 70B     & 8.14  & 0.228         \\
\hline
DeepSeek-R1 685B       & 11.1  & 0.0849        \\
\hline
DeepSeek-V3 685B       & 9.45  & 0.15          \\
\hline
\end{tabular}
\end{table}

\begin{figure}[H]
    \begin{center}
        \includegraphics[width=0.9\textwidth]{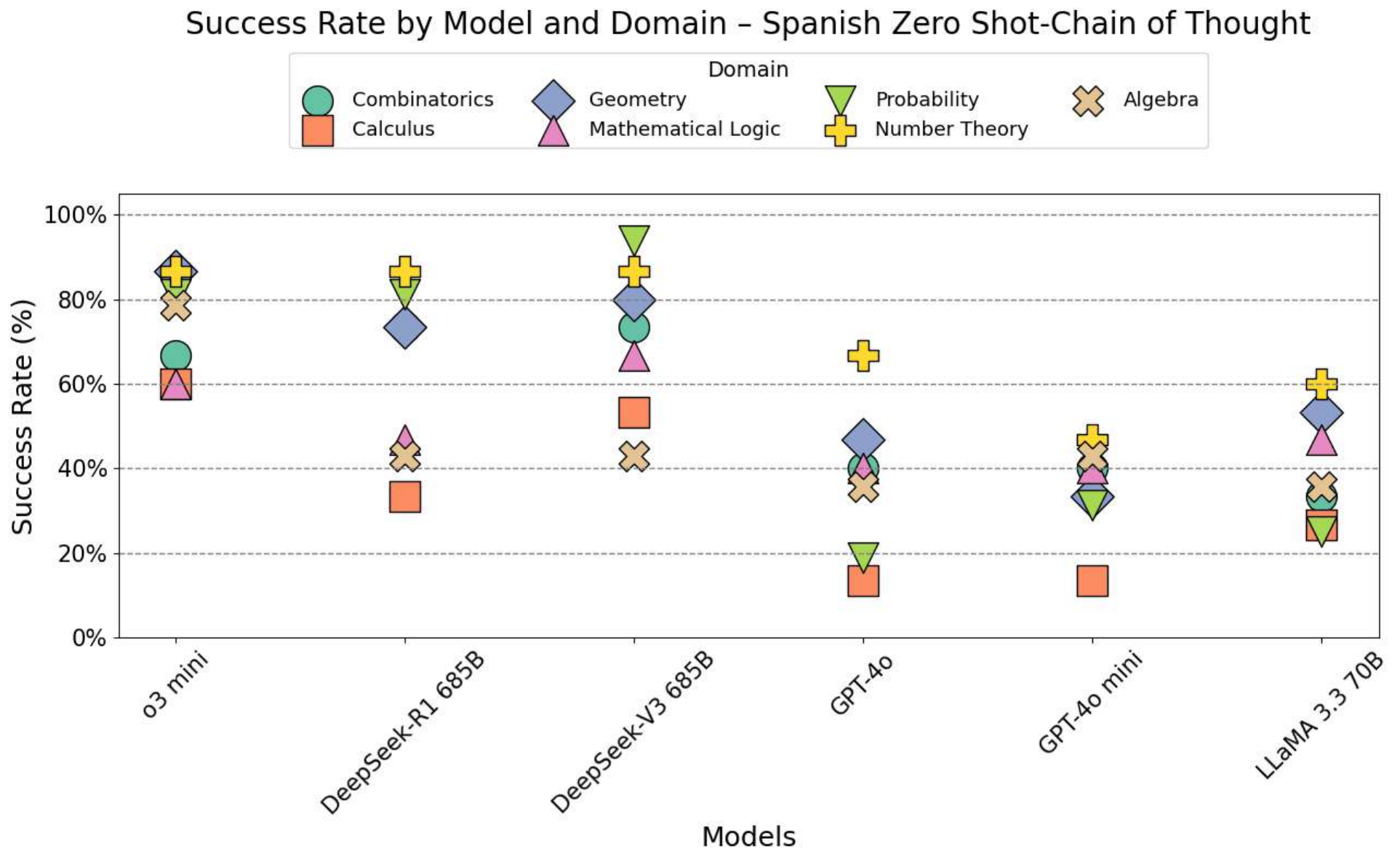}
    \end{center}
    \caption{The figure presents the success rate of large language models across different mathematical domains under the Spanish Zero-Shot Chain of Thought configuration. Each marker corresponds to a domain Combinatorics, Calculus, Geometry, Mathematical Logic, Probability, Number Theory, Algebra and shows the percentage of correctly solved problems by each model. Tab \ref{Table:domainChiSqauredSpanishCoTModels} and Tab. \ref{Tab:domainChiSquaredSpanishCoTdomain} report the statistical differences observed across models and mathematical domains.}\label{fig:domainChiSqauredSpanishCoT}
\end{figure}

In the Chain-of-Thought configuration (Fig. \ref{fig:domainChiSqauredSpanishCoT}, Tab. \ref{Table:domainChiSqauredSpanishCoTModels}), significant intra-model differences were detected for DeepSeek-R1 685B ($\chi^2$=17.1, df=6, p=0.00909) and DeepSeek-V3 685B ($\chi^2$=14.4, df=6, p=0.0259). Inter-model differences were also identified in Probability, Geometry, and Calculus (p$<$0.05) (Tab. \ref{Tab:domainChiSquaredSpanishCoTdomain}). These were the only models for which Chi-Squared tests reported significant performance variations across mathematical areas. Further post hoc analyses are detailed in the appendices (Appendix Tab. \ref{Tab:PosthoctestresultsforDeepSeek-R1bydomainundertheSpanish_ZS_CoTconfiguration.}, \ref{Tab:PosthoctestresultsforDeepSeek-R1bydomainundertheSpanish_ZS_CoTconfiguration.}, \ref{Tab:PosthoctestresultsforCalculusundertheSpanish_ZS_CoTconfiguration.}, \ref{Tab:PosthoctestresultsforGeometryundertheSpanish_ZS_CoTconfiguration.}, \ref{Tab:PosthoctestresultsforProbabilityundertheSpanish_ZS_CoTconfiguration.})

\begin{table}[H]
\centering
\caption{Chi-squared test results comparing performance across mathematical domains for each model under the Spanish Zero Shot Chain of Thought setting. Asterisks denote significance levels (* $p<0.05$, ** $p<0.01$, *** $p<0.001$).}\label{Table:domainChiSqauredSpanishCoTModels}
\begin{tabular}{|l|c|c|l|}
\hline
\textbf{Model} & \textbf{$\chi^2$ (df=6)} & \textbf{$p$-value}  \\
\hline
GPT-4o            & 12.3  & 0.0564         \\
\hline
GPT-4o mini       & 4.8   & 0.569          \\
\hline
o3 mini           & 6.61  & 0.358          \\
\hline
LLaMA 3.3 70B     & 6.88  & 0.332          \\
\hline
DeepSeek-R1 685B       & 17.1  & 0.00909       \textbf{**}  \\
\hline
DeepSeek-V3 685B       & 14.4  & 0.0259        \textbf{*}  \\
\hline
\end{tabular}
\end{table}

\begin{table}[H]
\centering
\caption{Post hoc chi-squared test results by domain – Spanish Zero Shot Chain of Thought. Asterisks denote significance levels (* $p<0.05$, ** $p<0.01$, *** $p<0.001$).}
\label{Tab:domainChiSquaredSpanishCoTdomain}
\begin{tabular}{|l|c|c|}
\hline
\textbf{Area} & \textbf{$\chi^2$ (df=5)} & \textbf{$p$-value} \\
\hline
Combinatorics      & 10.40 & 0.0637                    \\
\hline
Calculus           & 13.20 & 0.0216 *                 \\
\hline
Geometry           & 14.00 & 0.0157 *                 \\
\hline
Mathematical Logic &  3.60 & 0.6080                   \\
\hline
Probability        & 36.60 & $7.19 \times 10^{-7}$ *** \\
\hline
Number Theory      & 10.90 & 0.0532                   \\
\hline
Algebra            &  7.32 & 0.1980                   \\
\hline
\end{tabular}
\end{table}

As in English, Algebra, Number Theory, and Geometry were the top-performing domains for the strongest models. Probability showed more variance across models, particularly in Zero Shot-Chain of Thought. Performance gains from Chain of Thought Evaluation were consistent for o3-mini and partially for DeepSeek-R1 685B, while models like LLaMA 3.3 70B and GPT-4o mini showed less benefit or instability in reasoning under Spanish instructions.

\section{Discussion}

\textbf{Language-Based Performance Differences
}

\noindent Across languages, most models did not exhibit statistically significant performance differences between Spanish and English. McNemar tests confirmed that performance gaps due to prompt language were generally negligible. The notable exception was GPT-4o, which performed better on Spanish prompts in the zero-shot setting (53.3\% success in Spanish vs. 40.0\% in English, p = 0.008).This overall parity aligns with concerns raised by \citep{singh2024globalmmlu} and \citep{plaza2024}, who document that translation-based evaluations can introduce semantic drift and obscure model weaknesses in non-English settings. AI4Math addresses this by evaluating reasoning natively in Spanish, helping surface errors that English-centric benchmarks may miss. However, the GPT-4o result hints that certain models may leverage Spanish phrasing more effectively, underscoring the subtle influence of linguistic context on reasoning performance. Given our benchmark's modest size, we caution that broader conclusions about language invariance would benefit from expanded datasets and repeated evaluation.
\newpage 

\noindent \textbf{Domain-Specific Performance Challenges
}

\noindent Geometry, Combinatorics, and Probability emerged as the most challenging domains for all models, consistently yielding the lowest success rates. Even the top-performing models struggled with these problem types, indicating persistent weaknesses that transcend any single model. Qualitative analysis suggests that each of these domains demands capabilities current LLMs lack: Geometry often requires robust spatial reasoning (e.g., envisioning diagrams or transformations), Combinatorics involves complex case analysis and enumeration, and Probability questions rely on precise interpretation of linguistic cues and nested conditions. The persistence of errors in these areas across models and configurations (even under Zero-Shot CoT setting) suggests that mere scaling or prompt engineering yields diminishing returns here. These domains thus serve as reliable stress-tests and diagnostic indicators of a model’s reasoning limits in mathematics. Our findings align with prior benchmark evaluations: \citep{hendrycks2020mmlu} found that combinatorics and algebraic manipulation tasks remained particularly challenging in the MATH dataset, while \citep{glazer2024} highlighted persistent underperformance on symbolic and probabilistic reasoning in FrontierMath, even for frontier models.\newline 

\noindent \textbf{AI4Math as a Diagnostic Tool
}

\noindent AI4Math enables fine-grained analysis of model behavior by allowing researchers to examine topic- and language-specific error patterns. A model might perform well overall or in English prompts, yet fail systematically on Spanish inputs or particular mathematical topics. These insights support the growing emphasis on understanding why models fail, not just how often, and guide future improvements in model architecture and alignment.

\noindent \textbf{Surprising Model Rankings and Performance Parity
}

\noindent The leaderboard results on AI4Math challenge common expectations in the field. We found that the top models — o3-mini and DeepSeek-R1 685B — achieved success rates above 70\% in most configurations, with DeepSeek-V3 685B performing slightly lower but not significantly different. Statistical comparisons revealed no significant differences among their success rates, confirming that these three models performed comparably and far outpaced the remaining systems. Perhaps more surprisingly, these models outperformed several larger or more reputed LLMs, overturning assumptions that model size or name-brand equates to superior reasoning. For instance, o3-mini (a relatively compact, closed-source model) matched the performance of DeepSeek-R1 685B and even outscored OpenAI’s GPT-4o mini by a wide margin. In one configuration, o3-mini reached ~76\% in Spanish zero-shot, roughly double the success rate of GPT-4o mini. These outcomes echo the findings in Humanity’s Last Exam \citep{zheng2023humanits} , where model scale did not correlate consistently with reasoning quality. They also support arguments by \citep{solaiman2024impact} on the importance of third-party, independent evaluation to validate real-world performance. These reversals in expected ranking underscore the importance of domain-specific, independent evaluation. They demonstrate that smaller or open-source models, when fine-tuned or optimized for a task, can rival or exceed the performance of much larger models on complex mathematical reasoning. Such findings highlight that innovation is not monopolized by a few large institutions – strong models can emerge from diverse sources when tested under rigorous, transparent conditions. While DeepSeek-R1 685B and DeepSeek-V3 685B showed comparable performance, o3-mini achieved statistically significant improvements over models such as GPT-4o. This contrast illustrates the benchmark’s sensitivity to real performance gaps and its limitations in detecting subtle differences between closely matched models. Increasing dataset size and conducting repeated evaluations will help address this.

\noindent \textbf{Collaborative Benchmark Development
}

\noindent Our results also speak to the advantages of a collaborative, community-driven approach in benchmark creation. AI4Math was developed through a structured hackathon by Latin American STEM students, who authored and peer-reviewed 105 original university-level math problems in Spanish. This decentralized creation process ensured a high-quality, culturally relevant dataset while operating with minimal resources. The success of AI4Math demonstrates that rigorous benchmarks can be produced outside large institutions, aligning with calls for more inclusive and transparent evaluation frameworks. Community involvement and open peer review not only reduced development costs, but also injected diverse problem-solving perspectives, strengthening the benchmark’s coverage and fairness. We acknowledge, however, that this grassroots approach comes with challenges: maintaining consistency, achieving scale, and verifying solutions required significant human effort. Nonetheless, the AI4Math project underscores how collaborative methodologies can broaden participation in benchmark development, ultimately advancing evaluation practices alongside model capabilities.

\noindent \textbf{Limitations
}

\noindent Despite its contributions, AI4Math has several practical limitations that should temper the interpretation of our findings:

\begin{itemize}
    \item Sample Size: The benchmark contains 105 problems, which is modest compared to large-scale evaluations. This limited problem set restricts coverage of the vast space of mathematical tasks and may reduce statistical power in comparisons. We prioritized quality and originality of problems over quantity, but future work should expand the pool of questions for more robust generalization claims.
    \item Single Evaluation per Model: Due to resource and time constraints, we performed only one run per model per configuration, with no repeated trials or ensembling. Performance scores therefore do not capture variability across multiple attempts. A model’s true capabilities might be slightly under- or over-estimated from a single pass, and statistical tests could be underpowered. Repeated evaluations or ensemble methods could yield more stable estimates of model performance.
    \item Scope of Languages and Domains: Our evaluation was limited to Spanish (with English translations) in the domain of university mathematics. The results and conclusions are thus specific to this linguistic and subject context. It remains unclear how our findings extend to other natural languages or to different knowledge domains. In addition, we only tested zero-shot and zero-shot chain-of-thought configurations. Other prompting strategies, further languages, or interactive problem-solving scenarios were beyond our scope but represent important avenues for future investigation.

\end{itemize}

\noindent \textbf{Next Steps}

\noindent Future work will focus on four main areas. First, we will expand the benchmark with a larger and more diverse set of problems to improve statistical power and enable finer comparisons between close-performing models. Second, we plan to conduct repeated evaluations and prompt variations to better understand performance stability. Third, we aim to carry out a qualitative analysis of model-generated solutions to identify common reasoning patterns and failure modes. Finally, we will incorporate new and emerging models — including Grok-3 (xAI), Claude 3.7 Sonnet (Anthropic), and GPT-4.1 (OpenAI) — to ensure that AI4Math remains aligned with the latest developments in mathematical reasoning capabilities. These steps will help consolidate AI4Math as a scalable and diagnostic resource for evaluating mathematical reasoning in the next generation of language models.

\section{Conclusion}
AI4Math provides key insights and implications for multilingual evaluation and future benchmarks in mathematical reasoning. Most notably, our results highlight the need for benchmarking beyond English. A model’s problem-solving ability can vary across languages – even when the underlying math is the same – as seen in the subtle performance shifts on Spanish vs. English prompts. To build fair and generalizable AI, evaluation must reflect the linguistic diversity of real-world users, rather than relying solely on translated or English-only tests. Native-language benchmarks like AI4Math help ensure that progress in AI is measured on a broad, equitable basis.

At the same time, the persistent difficulties in Geometry, Combinatorics, and Probability across all models signal that current LLMs have likely hit a ceiling with scale and standard prompting in certain areas. Tackling these domains may require new techniques – for example, integrating visual-spatial reasoning, formal logic tools, or domain-specific modules – rather than expecting improvements from larger models alone. Future benchmark initiatives should consider incorporating such aspects, or designing hybrid evaluations, to drive progress on these stubborn challenges.

Finally, AI4Math illustrates the value of community-driven, diagnostic benchmarks that go beyond aggregate metrics to reveal nuanced model behavior. The unexpected performance gaps and the role of linguistic context underscore the need for open, detailed evaluation frameworks that evolve alongside language models. We hope this benchmark contributes to a more transparent and inclusive landscape for evaluating reasoning in multilingual AI. 

\noindent \textbf{The dataset is available upon request; please contact the corresponding author.}

\section*{Acknowledgments}
We thank the participants of the \textit{AI4MATH Challenge} hackathon for their enthusiasm and collaboration in developing the original mathematical problems: Everardo Urquiza Trejo, José Rubén Maldonado Herrera, Karla Sofía Zavala Álvarez, Jissela Johana Arcos Molina, Jesús Tadeo Cruz Soto, and Jaime Esteban Montenegro Barón.

\noindent We also thank the second-round reviewer Karla Sofía Zavala Álvarez for her valuable assistance in validating the problems. Ivanna Daniela Alvarado Morales and Monica Ulloa provided insightful suggestions on the writing and clarity of the final manuscript. Alan Ochoa González assisted significantly in running the open models. Special recognition goes to Jaime Sevilla for his substantial feedback throughout the project.

\bibliography{references}

\section{Complementary material}

\begin{table}[ht]
    \centering
    \caption{Success rate per domain under the English\_ZS configuration. The table shows the number of problems and percentage of correctly solved problems per model.}\label{Tab:Success_rate_per_domainundertheEnglishZSconfiguration}
    \resizebox{\textwidth}{!}{%
    \begin{tabular}{|l|c|c|c|c|c|c|c|}
    \hline
    \textbf{Area} & \textbf{N} & \textbf{GPT-4o} & \textbf{GPT-4o mini} & \textbf{o3 mini} & \textbf{LLaMA 3.3 70B} & \textbf{DeepSeek-R1 685B} & \textbf{DeepSeek-V3 685B} \\
    \hline
    Combinatorics     & 15 & 33.33\% & 26.67\% & 66.67\% & 26.67\% & 53.33\% & 40.00\% \\
    \hline
    Calculus          & 15 & 53.33\% & 46.67\% & 73.33\% & 53.33\% & 73.33\% & 53.33\% \\
    \hline
    Geometry          & 15 & 6.67\%  & 13.33\% & 80.00\% & 20.00\% & 46.67\% & 33.33\% \\
    \hline
    Mathematical Logic& 15 & 46.67\% & 46.67\% & 80.00\% & 53.33\% & 80.00\% & 73.33\% \\
    \hline
    Probability       & 16 & 31.25\% & 31.25\% & 56.25\% & 31.25\% & 62.50\% & 50.00\% \\
    \hline
    Number Theory     & 15 & 33.33\% & 26.67\% & 100.00\% & 33.33\% & 100.00\% & 100.00\% \\
    \hline
    Algebra           & 14 & 78.57\% & 57.14\% & 85.71\% & 50.00\% & 92.86\% & 92.86\% \\
    \hline
    \end{tabular}%
    }
\end{table}

\begin{table}[ht]
\centering
\caption{Success rate per domain under the English\_ZS\_CoT configuration. The table shows the number of problems and percentage of correctly solved problems per model.}\label{Tab:SuccessrateperdomainundertheEnglishZSconfiguration.}
\resizebox{\textwidth}{!}{%
\begin{tabular}{|l|c|c|c|c|c|c|c|}
\hline
\textbf{Area} & \textbf{N} & \textbf{GPT-4o} & \textbf{GPT-4o mini} & \textbf{o3 mini} & \textbf{LLaMA 3.3 70B} & \textbf{DeepSeek-R1 685B} & \textbf{DeepSeek-V3 685B} \\
\hline
Combinatorics     & 15 & 46.67\% & 40.00\% & 73.33\% & 33.33\% & 73.33\% & 46.67\% \\
\hline
Calculus          & 15 & 40.00\% & 40.00\% & 73.33\% & 33.33\% & 73.33\% & 60.00\% \\
\hline
Geometry          & 15 & 20.00\% & 26.67\% & 73.33\% & 20.00\% & 60.00\% & 46.67\% \\
\hline
Mathematical Logic& 15 & 60.00\% & 33.33\% & 80.00\% & 53.33\% & 80.00\% & 80.00\% \\
\hline
Probability       & 16 & 37.50\% & 31.25\% & 56.25\% & 31.25\% & 56.25\% & 50.00\% \\
\hline
Number Theory     & 15 & 40.00\% & 40.00\% & 93.33\% & 20.00\% & 100.00\% & 80.00\% \\
\hline
Algebra           & 14 & 78.57\% & 50.00\% & 78.57\% & 64.29\% & 92.86\% & 92.86\% \\
\hline
\end{tabular}%
}
\end{table}

\begin{table}[ht]
\centering
\caption{Success rate per domain under the Spanish\_ZS configuration. The table shows the number of problems and percentage of correctly solved problems per model.}\label{Tab:SuccessrateperdomainundertheSpanish_ZSconfiguration.}
\resizebox{\textwidth}{!}{%
\begin{tabular}{|l|c|c|c|c|c|c|c|}
\hline
\textbf{Area} & \textbf{N} & \textbf{GPT-4o} & \textbf{GPT-4o mini} & \textbf{o3 mini} & \textbf{LLaMA 3.3 70B} & \textbf{DeepSeek-R1 685B} & \textbf{DeepSeek-V3 685B} \\
\hline
Combinatorics     & 15 & 73.33\% & 33.33\% & 73.33\% & 40.00\% & 73.33\% & 73.33\% \\
\hline
Calculus          & 15 & 20.00\% & 13.33\% & 73.33\% & 20.00\% & 53.33\% & 46.67\% \\
\hline
Geometry          & 15 & 66.67\% & 33.33\% & 86.67\% & 40.00\% & 80.00\% & 80.00\% \\
\hline
Mathematical Logic& 15 & 53.33\% & 40.00\% & 60.00\% & 26.67\% & 53.33\% & 60.00\% \\
\hline
Probability       & 16 & 31.25\% & 31.25\% & 81.25\% & 12.50\% & 75.00\% & 87.50\% \\
\hline
Number Theory     & 15 & 80.00\% & 66.67\% & 86.67\% & 53.33\% & 93.33\% & 80.00\% \\
\hline
Algebra           & 14 & 50.00\% & 28.57\% & 71.43\% & 28.57\% & 50.00\% & 57.14\% \\
\hline
\end{tabular}%
}
\end{table}

\begin{table}[ht]
\centering
\caption{Success rate per domain under the Spanish\_ZS\_CoT configuration. The table shows the number of problems and percentage of correctly solved problems per model.}\label{Tab:SuccessrateperdomainundertheSpanish_ZS_CoTconfiguration.}
\resizebox{\textwidth}{!}{%
\begin{tabular}{|l|c|c|c|c|c|c|c|}
\hline
\textbf{Area} & \textbf{N} & \textbf{GPT-4o} & \textbf{GPT-4o mini} & \textbf{o3 mini} & \textbf{LLaMA 3.3 70B} & \textbf{DeepSeek-R1 685B} & \textbf{DeepSeek-V3 685B} \\
\hline
Combinatorics     & 15 & 40.00\% & 40.00\% & 66.67\% & 33.33\% & 73.33\% & 73.33\% \\
\hline
Calculus          & 15 & 13.33\% & 13.33\% & 60.00\% & 26.67\% & 33.33\% & 53.33\% \\
\hline
Geometry          & 15 & 46.67\% & 33.33\% & 86.67\% & 53.33\% & 73.33\% & 80.00\% \\
\hline
Mathematical Logic& 15 & 40.00\% & 40.00\% & 60.00\% & 46.67\% & 46.67\% & 66.67\% \\
\hline
Probability       & 16 & 18.75\% & 31.25\% & 81.25\% & 25.00\% & 81.25\% & 93.75\% \\
\hline
Number Theory     & 15 & 66.67\% & 46.67\% & 86.67\% & 60.00\% & 86.67\% & 86.67\% \\
\hline
Algebra           & 14 & 35.71\% & 42.86\% & 78.57\% & 35.71\% & 42.86\% & 42.86\% \\
\hline
\end{tabular}%
}
\end{table}

\begin{table}[ht]
    \centering
    \caption{Post hoc test results for English\_ZS configuration. Adjusted p-values and significance levels are reported.}
    \label{Tab:PosthoctestresultsforEnglish_ZSconfiguration}
    \resizebox{\textwidth}{!}{%
    \begin{tabular}{|l|l|c|c|c|}
    \hline
    \textbf{Group 1} & \textbf{Group 2} & \textbf{p} & \textbf{p.adj} & \textbf{p.adj.signif} \\
    \hline
    GPT-4o     & GPT-4o mini   & 5.69E-01 & 1.00E+00 & ns   \\
    GPT-4o     & o3 mini       & 1.02E-07 & 1.32E-06 & **** \\
    GPT-4o mini & o3 mini      & 2.22E-09 & 3.34E-08 & **** \\
    GPT-4o     & LLaMA         & 8.88E-01 & 1.00E+00 & ns   \\
    GPT-4o mini & LLaMA        & 7.75E-01 & 1.00E+00 & ns   \\
    o3 mini    & LLaMA         & 2.33E-08 & 3.26E-07 & **** \\
    GPT-4o     & DeepSeek-R1 685B   & 4.44E-06 & 4.44E-05 & **** \\
    GPT-4o mini & DeepSeek-R1 685B & 1.44E-07 & 1.73E-06 & **** \\
    o3 mini    & DeepSeek-R1 685B   & 5.25E-01 & 1.00E+00 & ns   \\
    LLaMA      & DeepSeek-R1 685B \\  
    GPT-4o     & DeepSeek-V3 685B   & 1.50E-03 & 1.05E-02 & *    \\
    GPT-4o mini & DeepSeek-V3 685B  & 1.11E-04 & 1.00E-03 & ***  \\
    o3 mini    & DeepSeek-V3 685B   & 3.50E-02 & 2.10E-01 & ns   \\
    LLaMA      & DeepSeek-V3 685B   & 5.60E-04 & 4.48E-03 & **   \\
    DeepSeek-R1 685B  & DeepSeek-V3 685B  & 1.84E-01 & 9.22E-01 & ns   \\
    \hline
    \end{tabular}%
    }
\end{table}

\begin{table}[ht]
\centering
\caption{Post hoc test results for English\_ZS\_CoT configuration. Adjusted p-values and significance levels are reported.}\label{Tab:PosthoctestresultsforEnglish_ZS_CoTconfiguration.}
\resizebox{\textwidth}{!}{%
\begin{tabular}{|l|l|c|c|c|}
\hline
\textbf{Group 1} & \textbf{Group 2} & \textbf{p} & \textbf{p.adj} & \textbf{p.adj.signif} \\
\hline
GPT-4o     & GPT-4o mini   & 2.62E-01 & 8.26E-01 & ns   \\
GPT-4o     & o3 mini       & 2.29E-05 & 2.29E-04 & ***  \\
GPT-4o mini & o3 mini      & 5.82E-08 & 6.98E-07 & **** \\
GPT-4o     & LLaMA         & 2.07E-01 & 8.26E-01 & ns   \\
GPT-4o mini & LLaMA        & 1.00E+00 & 1.00E+00 & ns   \\
o3 mini    & LLaMA         & 2.75E-08 & 3.57E-07 & **** \\
GPT-4o     & DeepSeek-R1 685B   & 1.16E-05 & 1.28E-04 & ***  \\
GPT-4o mini & DeepSeek-R1 685B  & 2.54E-08 & 3.56E-07 & **** \\
o3 mini    & DeepSeek-R1 685B   & 1.00E+00 & 1.00E+00 & ns   \\
LLaMA      & DeepSeek-R1 685B   & 1.18E-08 & 1.77E-07 & **** \\
GPT-4o     & DeepSeek-V3 685B   & 8.37E-03 & 5.86E-02 & ns   \\
GPT-4o mini & DeepSeek-V3 685B  & 1.11E-04 & 8.88E-04 & ***  \\
o3 mini    & DeepSeek-V3 685B   & 1.32E-01 & 6.61E-01 & ns   \\
LLaMA      & DeepSeek-V3 685B   & 6.27E-05 & 5.64E-04 & ***  \\
DeepSeek-R1 685B & DeepSeek-V3 685B  & 9.61E-02 & 5.77E-01 & ns   \\
\hline
\end{tabular}%
}
\end{table}

\begin{table}[ht]
\centering
\caption{Post hoc test results for Spanish\_ZS configuration. Adjusted p-values and significance levels are reported.}\label{Tab:PosthoctestresultsforSpanish_ZSconfiguration.}
\resizebox{\textwidth}{!}{%
\begin{tabular}{|l|l|c|c|c|}
\hline
\textbf{Group 1} & \textbf{Group 2} & \textbf{p} & \textbf{p.adj} & \textbf{p.adj.signif} \\
\hline
GPT-4o     & GPT-4o mini   & 1.24E-02 & 8.68E-02 & ns   \\
GPT-4o     & o3 mini       & 8.92E-04 & 8.03E-03 & **   \\
GPT-4o mini & o3 mini      & 5.39E-09 & 7.54E-08 & **** \\
GPT-4o     & LLaMA         & 2.13E-03 & 1.70E-02 & *    \\
GPT-4o mini & LLaMA        & 6.61E-01 & 1.00E+00 & ns   \\
o3 mini    & LLaMA         & 1.93E-10 & 2.89E-09 & **** \\
GPT-4o     & DeepSeek-R1 685B   & 2.33E-02 & 1.40E-01 & ns   \\
GPT-4o mini & DeepSeek-R1 685B  & 1.33E-06 & 1.46E-05 & **** \\
o3 mini    & DeepSeek-R1 685B   & 3.52E-01 & 1.00E+00 & ns   \\
LLaMA      & DeepSeek-R1 685B   & 7.34E-08 & 9.54E-07 & **** \\
GPT-4o     & DeepSeek-V3 685B   & 3.39E-02 & 1.69E-01 & ns   \\
GPT-4o mini & DeepSeek-V3 685B  & 2.65E-06 & 2.65E-05 & **** \\
o3 mini    & DeepSeek-V3 685B   & 2.80E-01 & 1.00E+00 & ns   \\
LLaMA      & DeepSeek-V3 685B   & 1.57E-07 & 1.88E-06 & **** \\
DeepSeek-R1 685B & DeepSeek-V3 685B  & 1.00E+00 & 1.00E+00 & ns   \\
\hline
\end{tabular}%
}
\end{table}

\begin{table}[ht]
\centering
\caption{Post hoc test results for Spanish\_ZS\_CoT configuration. Adjusted p-values and significance levels are reported.}\label{Tab:PosthoctestresultsforSpanish_ZS_CoTconfiguration.}
\resizebox{\textwidth}{!}{%
\begin{tabular}{|l|l|c|c|c|}
\hline
\textbf{Group 1} & \textbf{Group 2} & \textbf{p} & \textbf{p.adj} & \textbf{p.adj.signif} \\
\hline
GPT-4o     & GPT-4o mini   & 8.86E-01 & 1.00E+00 & ns   \\
GPT-4o     & o3 mini       & 1.30E-07 & 1.82E-06 & **** \\
GPT-4o mini & o3 mini      & 2.93E-08 & 4.39E-07 & **** \\
GPT-4o     & LLaMA         & 7.77E-01 & 1.00E+00 & ns   \\
GPT-4o mini & LLaMA        & 5.69E-01 & 1.00E+00 & ns   \\
o3 mini    & LLaMA         & 1.06E-06 & 1.27E-05 & **** \\
GPT-4o     & DeepSeek-R1 685B   & 1.25E-06 & 1.37E-05 & **** \\
GPT-4o mini & DeepSeek-R1 685B  & 3.09E-07 & 4.02E-06 & **** \\
o3 mini    & DeepSeek-R1 685B   & 7.56E-01 & 1.00E+00 & ns   \\
LLaMA      & DeepSeek-R1 685B   & 8.77E-06 & 8.77E-05 & **** \\
GPT-4o     & DeepSeek-V3 685B   & 3.33E-04 & 2.66E-03 & **   \\
GPT-4o mini & DeepSeek-V3 685B  & 1.11E-04 & 1.00E-03 & ***  \\
o3 mini    & DeepSeek-V3 685B   & 1.02E-01 & 6.12E-01 & ns   \\
LLaMA      & DeepSeek-V3 685B   & 1.50E-03 & 1.05E-02 & *    \\
DeepSeek-R1 685B & DeepSeek-V3 685B  & 2.40E-01 & 1.00E+00 & ns   \\
\hline
\end{tabular}%
}
\end{table}

\begin{table}[ht]
\centering
\caption{Post hoc test results for GPT-4o by domain under the English\_ZS configuration. Adjusted p-values and significance levels are reported.}\label{Tab:PosthoctestresultsforGPT-4obydomainundertheEnglish_ZSconfiguration.}
\resizebox{\textwidth}{!}{%
\begin{tabular}{|l|l|c|c|c|}
\hline
\textbf{Group 1} & \textbf{Group 2} & \textbf{p} & \textbf{p.adj} & \textbf{p.adj.signif} \\
\hline
Combinatorics       & Calculus           & 0.461   & 1       & ns   \\
Combinatorics       & Geometry           & 0.171   & 1       & ns   \\
Calculus            & Geometry           & 0.0168  & 0.337   & ns   \\
Combinatorics       & Mathematical logic & 0.709   & 1       & ns   \\
Calculus            & Mathematical logic & 1       & 1       & ns   \\
Geometry            & Mathematical logic & 0.039   & 0.685   & ns   \\
Combinatorics       & Probability        & 1       & 1       & ns   \\
Calculus            & Probability        & 0.378   & 1       & ns   \\
Geometry            & Probability        & 0.202   & 1       & ns   \\
Mathematical logic  & Probability        & 0.609   & 1       & ns   \\
Combinatorics       & Number Theory      & 1       & 1       & ns   \\
Calculus            & Number Theory      & 0.461   & 1       & ns   \\
Geometry            & Number Theory      & 0.171   & 1       & ns   \\
Mathematical logic  & Number Theory      & 0.709   & 1       & ns   \\
Probability         & Number Theory      & 1       & 1       & ns   \\
Combinatorics       & Algebra            & 0.0381  & 0.685   & ns   \\
Calculus            & Algebra            & 0.299   & 1       & ns   \\
Geometry            & Algebra            & 0.000383& 0.00805 & **   \\
Mathematical logic  & Algebra            & 0.166   & 1       & ns   \\
Probability         & Algebra            & 0.0261  & 0.495   & ns   \\
Number Theory       & Algebra            & 0.0381  & 0.685   & ns   \\
\hline
\end{tabular}%
}
\end{table}

\begin{table}[ht]
\centering
\caption{Post hoc test results for DeepSeek-R1 685B by domain under the English\_ZS configuration. Adjusted p-values and significance levels are reported.}\label{Tab:PosthoctestresultsforDeepSeek-R1bydomainundertheEnglish_ZSconfiguration.}
\resizebox{\textwidth}{!}{%
\begin{tabular}{|l|l|c|c|c|}
\hline
\textbf{Group 1} & \textbf{Group 2} & \textbf{p} & \textbf{p.adj} & \textbf{p.adj.signif} \\
\hline
Combinatorics       & Calculus           & 0.449    & 1       & ns \\
Combinatorics       & Geometry           & 1        & 1       & ns \\
Calculus            & Geometry           & 0.264    & 1       & ns \\
Combinatorics       & Mathematical logic & 0.245    & 1       & ns \\
Calculus            & Mathematical logic & 1        & 1       & ns \\
Geometry            & Mathematical logic & 0.130    & 1       & ns \\
Combinatorics       & Probability        & 0.879    & 1       & ns \\
Calculus            & Probability        & 0.795    & 1       & ns \\
Geometry            & Probability        & 0.600    & 1       & ns \\
Mathematical logic  & Probability        & 0.499    & 1       & ns \\
Combinatorics       & Number Theory      & 0.0096   & 0.192   & ns \\
Calculus            & Number Theory      & 0.107    & 1       & ns \\
Geometry            & Number Theory      & 0.00385  & 0.0809  & ns \\
Mathematical logic  & Number Theory      & 0.224    & 1       & ns \\
Probability         & Number Theory      & 0.0288   & 0.518   & ns \\
Combinatorics       & Algebra            & 0.0495   & 0.842   & ns \\
Calculus            & Algebra            & 0.369    & 1       & ns \\
Geometry            & Algebra            & 0.0223   & 0.424   & ns \\
Mathematical logic  & Algebra            & 0.642    & 1       & ns \\
Probability         & Algebra            & 0.126    & 1       & ns \\
Number Theory       & Algebra            & 0.972    & 1       & ns \\
\hline
\end{tabular}%
}
\end{table}

\begin{table}[ht]
\centering
\caption{Post hoc test results for DeepSeek-V3 685B by domain under the English\_ZS configuration. Adjusted p-values and significance levels are reported.}\label{Tab:PosthoctestresultsforDeepSeek-V3bydomainundertheEnglish_ZSconfiguration.}
\resizebox{\textwidth}{!}{%
\begin{tabular}{|l|l|c|c|c|}
\hline
\textbf{Group 1} & \textbf{Group 2} & \textbf{p} & \textbf{p.adj} & \textbf{p.adj.signif} \\
\hline
Combinatorics       & Calculus           & 0.714    & 1       & ns   \\
Combinatorics       & Geometry           & 1        & 1       & ns   \\
Calculus            & Geometry           & 0.461    & 1       & ns   \\
Combinatorics       & Mathematical logic & 0.141    & 1       & ns   \\
Calculus            & Mathematical logic & 0.449    & 1       & ns   \\
Geometry            & Mathematical logic & 0.0673   & 0.875   & ns   \\
Combinatorics       & Probability        & 0.843    & 1       & ns   \\
Calculus            & Probability        & 1        & 1       & ns   \\
Geometry            & Probability        & 0.565    & 1       & ns   \\
Mathematical logic  & Probability        & 0.335    & 1       & ns   \\
Combinatorics       & Number Theory      & 0.00144  & 0.0287  & *    \\
Calculus            & Number Theory      & 0.0096   & 0.158   & ns   \\
Geometry            & Number Theory      & 0.000491 & 0.0103  & *    \\
Mathematical logic  & Number Theory      & 0.107    & 1       & ns   \\
Probability         & Number Theory      & 0.00563  & 0.101   & ns   \\
Combinatorics       & Algebra            & 0.00928  & 0.158   & ns   \\
Calculus            & Algebra            & 0.0495   & 0.694   & ns   \\
Geometry            & Algebra            & 0.00352  & 0.0669  & ns   \\
Mathematical logic  & Algebra            & 0.369    & 1       & ns   \\
Probability         & Algebra            & 0.0311   & 0.466   & ns   \\
Number Theory       & Algebra            & 0.972    & 1       & ns   \\
\hline
\end{tabular}%
}
\end{table}

\begin{table}[ht]
\centering
\caption{Post hoc test results for DeepSeek-R1 by domain under the English\_ZS\_CoT configuration. Adjusted p-values and significance levels are reported.}\label{Tab:PosthoctestresultsforDeepSeek-R1bydomainundertheEnglish_ZS_CoTconfiguration.}
\resizebox{\textwidth}{!}{%
\begin{tabular}{|l|l|c|c|c|}
\hline
\textbf{Group 1} & \textbf{Group 2} & \textbf{p} & \textbf{p.adj} & \textbf{p.adj.signif} \\
\hline
Combinatorics       & Calculus           & 1        & 1       & ns   \\
Combinatorics       & Geometry           & 0.699    & 1       & ns   \\
Calculus            & Geometry           & 0.699    & 1       & ns   \\
Combinatorics       & Mathematical logic & 1        & 1       & ns   \\
Calculus            & Mathematical logic & 1        & 1       & ns   \\
Geometry            & Mathematical logic & 0.426    & 1       & ns   \\
Combinatorics       & Probability        & 0.537    & 1       & ns   \\
Calculus            & Probability        & 0.537    & 1       & ns   \\
Geometry            & Probability        & 1        & 1       & ns   \\
Mathematical logic  & Probability        & 0.303    & 1       & ns   \\
Combinatorics       & Number Theory      & 0.107    & 1       & ns   \\
Calculus            & Number Theory      & 0.107    & 1       & ns   \\
Geometry            & Number Theory      & 0.0225   & 0.450   & ns   \\
Mathematical logic  & Number Theory      & 0.224    & 1       & ns   \\
Probability         & Number Theory      & 0.0131   & 0.275   & ns   \\
Combinatorics       & Algebra            & 0.369    & 1       & ns   \\
Calculus            & Algebra            & 0.369    & 1       & ns   \\
Geometry            & Algebra            & 0.103    & 1       & ns   \\
Mathematical logic  & Algebra            & 0.642    & 1       & ns   \\
Probability         & Algebra            & 0.0646   & 1       & ns   \\
Number Theory       & Algebra            & 0.972    & 1       & ns   \\
\hline
\end{tabular}%
}
\end{table}

\begin{table}[ht]
\centering
\caption{Post hoc test results for DeepSeek-V3 685B by domain under the English\_ZS\_CoT configuration. Adjusted p-values and significance levels are reported.}\label{Tab:PosthoctestresultsforDeepSeek-V3bydomainundertheEnglish_ZS_CoTconfiguration.}
\resizebox{\textwidth}{!}{%
\begin{tabular}{|l|l|c|c|c|}
\hline
\textbf{Group 1} & \textbf{Group 2} & \textbf{p} & \textbf{p.adj} & \textbf{p.adj.signif} \\
\hline
Combinatorics       & Calculus           & 0.714    & 1       & ns   \\
Combinatorics       & Geometry           & 1        & 1       & ns   \\
Calculus            & Geometry           & 0.714    & 1       & ns   \\
Combinatorics       & Mathematical logic & 0.130    & 1       & ns   \\
Calculus            & Mathematical logic & 0.426    & 1       & ns   \\
Geometry            & Mathematical logic & 0.130    & 1       & ns   \\
Combinatorics       & Probability        & 1        & 1       & ns   \\
Calculus            & Probability        & 0.843    & 1       & ns   \\
Geometry            & Probability        & 1        & 1       & ns   \\
Mathematical logic  & Probability        & 0.171    & 1       & ns   \\
Combinatorics       & Number Theory      & 0.130    & 1       & ns   \\
Calculus            & Number Theory      & 0.426    & 1       & ns   \\
Geometry            & Number Theory      & 0.130    & 1       & ns   \\
Mathematical logic  & Number Theory      & 1        & 1       & ns   \\
Probability         & Number Theory      & 0.171    & 1       & ns   \\
Combinatorics       & Algebra            & 0.0223   & 0.468   & ns   \\
Calculus            & Algebra            & 0.103    & 1       & ns   \\
Geometry            & Algebra            & 0.0223   & 0.468   & ns   \\
Mathematical logic  & Algebra            & 0.642    & 1       & ns   \\
Probability         & Algebra            & 0.0311   & 0.590   & ns   \\
Number Theory       & Algebra            & 0.642    & 1       & ns   \\
\hline
\end{tabular}%
}
\end{table}

\begin{table}[ht]
\centering
\caption{Post hoc test results for GPT-4o by domain under the Spanish\_ZS configuration. Adjusted p-values and significance levels are reported.}\label{Tab:PosthoctestresultsforGPT-4obydomainundertheSpanish_ZSconfiguration.}
\resizebox{\textwidth}{!}{%
\begin{tabular}{|l|l|c|c|c|}
\hline
\textbf{Group 1} & \textbf{Group 2} & \textbf{p} & \textbf{p.adj} & \textbf{p.adj.signif} \\
\hline
Combinatorics       & Calculus           & 0.0104   & 0.208   & ns \\
Combinatorics       & Geometry           & 1        & 1       & ns \\
Calculus            & Geometry           & 0.0271   & 0.487   & ns \\
Combinatorics       & Mathematical logic & 0.449    & 1       & ns \\
Calculus            & Mathematical logic & 0.13     & 1       & ns \\
Geometry            & Mathematical logic & 0.709    & 1       & ns \\
Combinatorics       & Probability        & 0.0473   & 0.804   & ns \\
Calculus            & Probability        & 0.761    & 1       & ns \\
Geometry            & Probability        & 0.107    & 1       & ns \\
Mathematical logic  & Probability        & 0.378    & 1       & ns \\
Combinatorics       & Number Theory      & 1        & 1       & ns \\
Calculus            & Number Theory      & 0.00349  & 0.0732  & ns \\
Geometry            & Number Theory      & 0.68     & 1       & ns \\
Mathematical logic  & Number Theory      & 0.245    & 1       & ns \\
Probability         & Number Theory      & 0.0181   & 0.343   & ns \\
Combinatorics       & Algebra            & 0.362    & 1       & ns \\
Calculus            & Algebra            & 0.191    & 1       & ns \\
Geometry            & Algebra            & 0.594    & 1       & ns \\
Mathematical logic  & Algebra            & 1        & 1       & ns \\
Probability         & Algebra            & 0.501    & 1       & ns \\
Number Theory       & Algebra            & 0.191    & 1       & ns \\
\hline
\end{tabular}%
}
\end{table}

\begin{table}[ht]
\centering
\caption{Post hoc test results for DeepSeek-R1 685B by domain under the Spanish\_ZS\_CoT configuration. Adjusted p-values and significance levels are reported.}\label{Tab:PosthoctestresultsforDeepSeek-R1bydomainundertheSpanish_ZS_CoTconfiguration.}
\resizebox{\textwidth}{!}{%
\begin{tabular}{|l|l|c|c|c|}
\hline
\textbf{Group 1} & \textbf{Group 2} & \textbf{p} & \textbf{p.adj} & \textbf{p.adj.signif} \\
\hline
Combinatorics       & Calculus           & 0.0673   & 1       & ns \\
Combinatorics       & Geometry           & 1        & 1       & ns \\
Calculus            & Geometry           & 0.0673   & 1       & ns \\
Combinatorics       & Mathematical logic & 0.264    & 1       & ns \\
Calculus            & Mathematical logic & 0.709    & 1       & ns \\
Geometry            & Mathematical logic & 0.264    & 1       & ns \\
Combinatorics       & Probability        & 0.923    & 1       & ns \\
Calculus            & Probability        & 0.0194   & 0.388   & ns \\
Geometry            & Probability        & 0.923    & 1       & ns \\
Mathematical logic  & Probability        & 0.102    & 1       & ns \\
Combinatorics       & Number Theory      & 0.648    & 1       & ns \\
Calculus            & Number Theory      & 0.00909  & 0.191   & ns \\
Geometry            & Number Theory      & 0.648    & 1       & ns \\
Mathematical logic  & Number Theory      & 0.0528   & 0.951   & ns \\
Probability         & Number Theory      & 1        & 1       & ns \\
Combinatorics       & Algebra            & 0.198    & 1       & ns \\
Calculus            & Algebra            & 0.885    & 1       & ns \\
Geometry            & Algebra            & 0.198    & 1       & ns \\
Mathematical logic  & Algebra            & 1        & 1       & ns \\
Probability         & Algebra            & 0.0723   & 1       & ns \\
Number Theory       & Algebra            & 0.0367   & 0.697   & ns \\
\hline
\end{tabular}%
}
\end{table}

\begin{table}[ht]
\centering
\caption{Post hoc test results for DeepSeek-V3 685B by domain under the Spanish\_ZS\_CoT configuration. Adjusted p-values and significance levels are reported.}\label{Tab:PosthoctestresultsforDeepSeek-V3bydomainundertheSpanish_ZS_CoTconfiguration.}
\resizebox{\textwidth}{!}{%
\begin{tabular}{|l|l|c|c|c|}
\hline
\textbf{Group 1} & \textbf{Group 2} & \textbf{p} & \textbf{p.adj} & \textbf{p.adj.signif} \\
\hline
Combinatorics       & Calculus           & 0.449    & 1       & ns \\
Combinatorics       & Geometry           & 1        & 1       & ns \\
Calculus            & Geometry           & 0.245    & 1       & ns \\
Combinatorics       & Mathematical logic & 1        & 1       & ns \\
Calculus            & Mathematical logic & 0.709    & 1       & ns \\
Geometry            & Mathematical logic & 0.680    & 1       & ns \\
Combinatorics       & Probability        & 0.291    & 1       & ns \\
Calculus            & Probability        & 0.0308   & 0.616   & ns \\
Geometry            & Probability        & 0.545    & 1       & ns \\
Mathematical logic  & Probability        & 0.146    & 1       & ns \\
Combinatorics       & Number Theory      & 0.648    & 1       & ns \\
Calculus            & Number Theory      & 0.111    & 1       & ns \\
Geometry            & Number Theory      & 1        & 1       & ns \\
Mathematical logic  & Number Theory      & 0.388    & 1       & ns \\
Probability         & Number Theory      & 0.953    & 1       & ns \\
Combinatorics       & Algebra            & 0.198    & 1       & ns \\
Calculus            & Algebra            & 0.847    & 1       & ns \\
Geometry            & Algebra            & 0.0935   & 1       & ns \\
Mathematical logic  & Algebra            & 0.360    & 1       & ns \\
Probability         & Algebra            & 0.0084   & 0.176   & ns \\
Number Theory       & Algebra            & 0.0367   & 0.697   & ns \\
\hline
\end{tabular}%
}
\end{table}

\begin{table}[ht]
\centering
\caption{Post hoc test results for Algebra under the English\_ZS configuration. Adjusted p-values and significance levels are reported.}\label{Tab:PosthoctestresultsforAlgebraundertheEnglish_ZSconfiguration.}
\resizebox{\textwidth}{!}{%
\begin{tabular}{|l|l|c|c|c|}
\hline
\textbf{Group 1} & \textbf{Group 2} & \textbf{p} & \textbf{p.adj} & \textbf{p.adj.signif} \\
\hline
GPT\_4o          & GPT\_4o\_mini       & 0.418   & 1       & ns \\
GPT\_4o          & o3\_mini            & 1       & 1       & ns \\
GPT\_4o\_mini     & o3\_mini            & 0.209   & 1       & ns \\
GPT\_4o          & LLaMA\_3\_3\_70B     & 0.237   & 1       & ns \\
GPT\_4o\_mini     & LLaMA\_3\_3\_70B     & 1       & 1       & ns \\
o3\_mini         & LLaMA\_3\_3\_70B     & 0.106   & 1       & ns \\
GPT\_4o          & DeepSeek\_R1 685B        & 0.589   & 1       & ns \\
GPT\_4o\_mini     & DeepSeek\_R1 685B       & 0.0809  & 1       & ns \\
o3\_mini         & DeepSeek\_R1 685B       & 1       & 1       & ns \\
LLaMA\_3\_3\_70B  & DeepSeek\_R1 685B        & 0.0365  & 0.547   & ns \\
GPT\_4o          & DeepSeek\_V3 685B        & 0.589   & 1       & ns \\
GPT\_4o\_mini     & DeepSeek\_V3 685B        & 0.0809  & 1       & ns \\
o3\_mini         & DeepSeek\_V3 685B        & 1       & 1       & ns \\
LLaMA\_3\_3\_70B  & DeepSeek\_V3 685B        & 0.0365  & 0.547   & ns \\
DeepSeek\_R1 685B     & DeepSeek\_V3 685B        & 1       & 1       & ns \\
\hline
\end{tabular}%
}
\end{table}

\begin{table}[ht]
\centering
\caption{Post hoc test results for Number Theory under the English\_ZS configuration. Adjusted p-values and significance levels are reported.}\label{Tab:PosthoctestresultsforNumberTheoryundertheEnglish_ZSconfiguration.}
\resizebox{\textwidth}{!}{%
\begin{tabular}{|l|l|c|c|c|}
\hline
\textbf{Group 1} & \textbf{Group 2} & \textbf{p} & \textbf{p.adj} & \textbf{p.adj.signif} \\
\hline
GPT\_4o          & GPT\_4o\_mini       & 1         & 1         & ns   \\
GPT\_4o          & o3\_mini            & 0.000491  & 0.00442   & **   \\
GPT\_4o\_mini     & o3\_mini            & 0.000151  & 0.00182   & **   \\
GPT\_4o          & LLaMA\_3\_3\_70B     & 1         & 1         & ns   \\
GPT\_4o\_mini     & LLaMA\_3\_3\_70B     & 1         & 1         & ns   \\
o3\_mini         & LLaMA\_3\_3\_70B     & 0.000491  & 0.00442   & **   \\
GPT\_4o          & DeepSeek\_R1 685B        & 0.000491  & 0.00442   & **   \\
GPT\_4o\_mini     & DeepSeek\_R1 685B        & 0.000151  & 0.00182   & **   \\
LLaMA\_3\_3\_70B  & DeepSeek\_R1 685B        & 0.000491  & 0.00442   & **   \\
GPT\_4o          & DeepSeek\_V3 685B        & 0.000491  & 0.00442   & **   \\
GPT\_4o\_mini     & DeepSeek\_V3 685B        & 0.000151  & 0.00182   & **   \\
LLaMA\_3\_3\_70B  & DeepSeek\_V3 685B        & 0.000491  & 0.00442   & **   \\
\hline
\end{tabular}%
}
\end{table}

\begin{table}[ht]
\centering
\caption{Post hoc test results for Geometry under the English\_ZS configuration. Adjusted p-values and significance levels are reported.}\label{Tab:PosthoctestresultsforGeometryundertheEnglish_ZSconfiguration.}
\resizebox{\textwidth}{!}{%
\begin{tabular}{|l|l|c|c|c|}
\hline
\textbf{Group 1} & \textbf{Group 2} & \textbf{p} & \textbf{p.adj} & \textbf{p.adj.signif} \\
\hline
GPT\_4o          & GPT\_4o\_mini       & 1         & 1         & ns   \\
GPT\_4o          & o3\_mini            & 0.000229  & 0.00344   & **   \\
GPT\_4o\_mini     & o3\_mini            & 0.000989  & 0.0138    & *    \\
GPT\_4o          & LLaMA\_3\_3\_70B     & 0.591     & 1         & ns   \\
GPT\_4o\_mini     & LLaMA\_3\_3\_70B     & 1         & 1         & ns   \\
o3\_mini         & LLaMA\_3\_3\_70B     & 0.00349   & 0.0453    & *    \\
GPT\_4o          & DeepSeek\_R1 685B        & 0.039     & 0.429     & ns   \\
GPT\_4o\_mini     & DeepSeek\_R1 685B       & 0.111     & 1         & ns   \\
o3\_mini         & DeepSeek\_R1 685B        & 0.13      & 1         & ns   \\
LLaMA\_3\_3\_70B  & DeepSeek\_R1 685B        & 0.245     & 1         & ns   \\
GPT\_4o          & DeepSeek\_V3 685B        & 0.171     & 1         & ns   \\
GPT\_4o\_mini     & DeepSeek\_V3 685B        & 0.388     & 1         & ns   \\
o3\_mini         & DeepSeek\_V3 685B        & 0.0271    & 0.325     & ns   \\
LLaMA\_3\_3\_70B  & DeepSeek\_V3 685B        & 0.68      & 1         & ns   \\
DeepSeek\_R1 685B     & DeepSeek\_V3 685B        & 0.709     & 1         & ns   \\
\hline
\end{tabular}%
}
\end{table}

\begin{table}[ht]
\centering
\caption{Post hoc test results for Number Theory under the English\textunderscore ZS\textunderscore CoT configuration. Adjusted p-values and significance levels are reported.}
\label{tab:posthoc_number_theory_english_zs_cot}
\resizebox{\textwidth}{!}{%
\begin{tabular}{|l|l|c|c|c|}
\hline
\textbf{Group 1} & \textbf{Group 2} & \textbf{p} & \textbf{p.adj} & \textbf{p.adj.signif} \\
\hline
GPT\_4o          & GPT\_4o\_mini       & 1.00E+00  & 1       & ns   \\
GPT\_4o          & o3\_mini            & 6.71E-03  & 0.0671  & ns   \\
GPT\_4o\_mini     & o3\_mini            & 6.71E-03  & 0.0671  & ns   \\
GPT\_4o          & LLaMA               & 4.26E-01  & 1       & ns   \\
GPT\_4o\_mini     & LLaMA               & 4.26E-01  & 1       & ns   \\
o3\_mini         & LLaMA               & 2.29E-04  & 0.00321 & **   \\
GPT\_4o          & DeepSeek\_R1 685B        & 1.44E-03  & 0.0187  & *    \\
GPT\_4o\_mini     & DeepSeek\_R1 685B        & 1.44E-03  & 0.0187  & *    \\
o3\_mini         & DeepSeek\_R1 685B        & 1.00E+00  & 1       & ns   \\
LLaMA           & DeepSeek\_R1 685B        & 4.14E-05  & 0.000621& ***  \\
GPT\_4o          & DeepSeek\_V3 685B        & 6.24E-02  & 0.499   & ns   \\
GPT\_4o\_mini     & DeepSeek\_V3 685B        & 6.24E-02  & 0.499   & ns   \\
o3\_mini         & DeepSeek\_V3 685B        & 5.91E-01  & 1       & ns   \\
LLaMA           & DeepSeek\_V3 685B        & 3.49E-03  & 0.0384  & *    \\
DeepSeek\_R1 685B     & DeepSeek\_V3 685B        & 2.24E-01  & 1       & ns   \\
\hline
\end{tabular}%
}
\end{table}

\begin{table}[ht]
\centering
\caption{Post hoc test results for Mathematical Logic under the English\textunderscore ZS\textunderscore CoT configuration. Adjusted p-values and significance levels are reported.}
\label{tab:posthoc_mathlogic_english_zs_cot}
\resizebox{\textwidth}{!}{%
\begin{tabular}{|l|l|c|c|c|}
\hline
\textbf{Group 1} & \textbf{Group 2} & \textbf{p} & \textbf{p.adj} & \textbf{p.adj.signif} \\
\hline
GPT\_4o          & GPT\_4o\_mini       & 0.272     & 1       & ns \\
GPT\_4o          & o3\_mini            & 0.426     & 1       & ns \\
GPT\_4o\_mini     & o3\_mini            & 0.0271    & 0.406   & ns \\
GPT\_4o          & LLaMA               & 1         & 1       & ns \\
GPT\_4o\_mini     & LLaMA               & 0.461     & 1       & ns \\
o3\_mini         & LLaMA               & 0.245     & 1       & ns \\
GPT\_4o          & DeepSeek\_R1 685B   & 0.426     & 1       & ns \\
GPT\_4o\_mini     & DeepSeek\_R1 685B   & 0.0271    & 0.406   & ns \\
o3\_mini         & DeepSeek\_R1 685B   & 1         & 1       & ns \\
LLaMA           & DeepSeek\_R1 685B   & 0.245     & 1       & ns \\
GPT\_4o          & DeepSeek\_V3 685B   & 0.426     & 1       & ns \\
GPT\_4o\_mini     & DeepSeek\_V3 685B   & 0.0271    & 0.406   & ns \\
o3\_mini         & DeepSeek\_V3 685B   & 1         & 1       & ns \\
LLaMA           & DeepSeek\_V3 685B   & 0.245     & 1       & ns \\
DeepSeek\_R1 685B & DeepSeek\_V3 685B  & 1         & 1       & ns \\
\hline
\end{tabular}%
}
\end{table}

\begin{table}[ht]
\centering
\caption{Post hoc test results for Geometry under the English\textunderscore ZS\textunderscore CoT configuration. Adjusted p-values and significance levels are reported.}
\label{tab:posthoc_geometry_english_zs_cot}
\resizebox{\textwidth}{!}{%
\begin{tabular}{|l|l|c|c|c|}
\hline
\textbf{Group 1} & \textbf{Group 2} & \textbf{p} & \textbf{p.adj} & \textbf{p.adj.signif} \\
\hline
GPT\_4o          & GPT\_4o\_mini       & 1         & 1       & ns \\
GPT\_4o          & o3\_mini            & 0.0104    & 0.156   & ns \\
GPT\_4o\_mini     & o3\_mini            & 0.0285    & 0.370   & ns \\
GPT\_4o          & LLaMA               & 1         & 1       & ns \\
GPT\_4o\_mini     & LLaMA               & 1         & 1       & ns \\
o3\_mini         & LLaMA               & 0.0104    & 0.156   & ns \\
GPT\_4o          & DeepSeek\_R1 685B   & 0.0624    & 0.749   & ns \\
GPT\_4o\_mini     & DeepSeek\_R1 685B   & 0.141     & 1       & ns \\
o3\_mini         & DeepSeek\_R1 685B   & 0.699     & 1       & ns \\
LLaMA           & DeepSeek\_R1 685B   & 0.0624    & 0.749   & ns \\
GPT\_4o          & DeepSeek\_V3 685B   & 0.245     & 1       & ns \\
GPT\_4o\_mini     & DeepSeek\_V3 685B   & 0.449     & 1       & ns \\
o3\_mini         & DeepSeek\_V3 685B   & 0.264     & 1       & ns \\
LLaMA           & DeepSeek\_V3 685B   & 0.245     & 1       & ns \\
DeepSeek\_R1 685B & DeepSeek\_V3 685B  & 0.714     & 1       & ns \\
\hline
\end{tabular}%
}
\end{table}

\begin{table}[ht]
\centering
\caption{Post hoc test results for Probability under the Spanish\textunderscore ZS configuration. Adjusted p-values and significance levels are reported.}
\label{tab:posthoc_probability_spanish_zs}
\resizebox{\textwidth}{!}{%
\begin{tabular}{|l|l|c|c|c|}
\hline
\textbf{Group 1} & \textbf{Group 2} & \textbf{p} & \textbf{p.adj} & \textbf{p.adj.signif} \\
\hline
GPT\_4o          & GPT\_4o\_mini       & 1         & 1         & ns   \\
GPT\_4o          & o3\_mini            & 0.0126    & 0.126     & ns   \\
GPT\_4o\_mini     & o3\_mini            & 0.0126    & 0.126     & ns   \\
GPT\_4o          & LLaMA\_3\_3\_70B     & 0.392     & 1         & ns   \\
GPT\_4o\_mini     & LLaMA\_3\_3\_70B     & 0.392     & 1         & ns   \\
o3\_mini         & LLaMA\_3\_3\_70B     & 0.000396  & 0.00555   & **   \\
GPT\_4o          & DeepSeek\_R1 685B   & 0.0335    & 0.268     & ns   \\
GPT\_4o\_mini     & DeepSeek\_R1 685B   & 0.0335    & 0.268     & ns   \\
o3\_mini         & DeepSeek\_R1 685B   & 1         & 1         & ns   \\
LLaMA\_3\_3\_70B  & DeepSeek\_R1 685B   & 0.00134   & 0.0174    & *    \\
GPT\_4o          & DeepSeek\_V3 685B   & 0.00398   & 0.0478    & *    \\
GPT\_4o\_mini     & DeepSeek\_V3 685B   & 0.00398   & 0.0478    & *    \\
o3\_mini         & DeepSeek\_V3 685B   & 1         & 1         & ns   \\
LLaMA\_3\_3\_70B  & DeepSeek\_V3 685B   & 0.000101  & 0.00151   & **   \\
DeepSeek\_R1 685B & DeepSeek\_V3 685B  & 0.651     & 1         & ns   \\
\hline
\end{tabular}%
}
\end{table}

\begin{table}[ht]
\centering
\caption{Post hoc test results for Geometry under the Spanish ZS configuration. Adjusted p-values and significance levels are reported.}
\label{tab:posthoc_geometry_spanish_zs}
\resizebox{\textwidth}{!}{%
\begin{tabular}{|l|l|c|c|c|}
\hline
\textbf{Group 1} & \textbf{Group 2} & \textbf{p} & \textbf{p.adj} & \textbf{p.adj.signif} \\
\hline
GPT\_4o          & GPT\_4o\_mini       & 0.144     & 1       & ns \\
GPT\_4o          & o3\_mini            & 0.388     & 1       & ns \\
GPT\_4o\_mini    & o3\_mini            & 0.00909   & 0.136   & ns \\
GPT\_4o          & LLaMA\_3\_3\_70B    & 0.272     & 1       & ns \\
GPT\_4o\_mini    & LLaMA\_3\_3\_70B    & 1         & 1       & ns \\
o3\_mini         & LLaMA\_3\_3\_70B    & 0.023     & 0.322   & ns \\
GPT\_4o          & DeepSeek\_R1 685B   & 0.680     & 1       & ns \\
GPT\_4o\_mini    & DeepSeek\_R1 685B   & 0.0271    & 0.352   & ns \\
o3\_mini         & DeepSeek\_R1 685B   & 1         & 1       & ns \\
LLaMA\_3\_3\_70B & DeepSeek\_R1 685B   & 0.0624    & 0.686   & ns \\
GPT\_4o          & DeepSeek\_V3 685B   & 0.680     & 1       & ns \\
GPT\_4o\_mini    & DeepSeek\_V3 685B   & 0.0271    & 0.352   & ns \\
o3\_mini         & DeepSeek\_V3 685B   & 1         & 1       & ns \\
LLaMA\_3\_3\_70B & DeepSeek\_V3 685B   & 0.0624    & 0.686   & ns \\
DeepSeek\_R1 685B & DeepSeek\_V3 685B  & 1         & 1       & ns \\
\hline
\end{tabular}%
}
\end{table}

\begin{table}[ht]
\centering
\caption{Post hoc test results for Calculus under the Spanish\_ZS configuration. Adjusted p-values and significance levels are reported.}\label{Tab:PosthoctestresultsforCalculusundertheSpanish_ZSconfiguration.}
\resizebox{\textwidth}{!}{%
\begin{tabular}{|l|l|c|c|c|}
\hline
\textbf{Group 1} & \textbf{Group 2} & \textbf{p} & \textbf{p.adj} & \textbf{p.adj.signif} \\
\hline
GPT\_4o          & GPT\_4o\_mini       & 1         & 1       & ns \\
GPT\_4o          & o3\_mini            & 0.0104    & 0.146   & ns \\
GPT\_4o\_mini     & o3\_mini            & 0.0032    & 0.0481  & *  \\
GPT\_4o          & LLaMA\_3\_3\_70B     & 1         & 1       & ns \\
GPT\_4o\_mini     & LLaMA\_3\_3\_70B     & 1         & 1       & ns \\
o3\_mini         & LLaMA\_3\_3\_70B     & 0.0104    & 0.146   & ns \\
GPT\_4o          & DeepSeek\_R1 685B        & 0.13      & 1       & ns \\
GPT\_4o\_mini     & DeepSeek\_R1 685B        & 0.0528    & 0.634   & ns \\
o3\_mini         & DeepSeek\_R1 685B        & 0.449     & 1       & ns \\
LLaMA\_3\_3\_70B  & DeepSeek\_R1 685B        & 0.13      & 1       & ns \\
GPT\_4o          & DeepSeek\_V3 685B        & 0.245     & 1       & ns \\
GPT\_4o\_mini     & DeepSeek\_V3 685B        & 0.111     & 1       & ns \\
o3\_mini         & DeepSeek\_V3 685B        & 0.264     & 1       & ns \\
LLaMA\_3\_3\_70B  & DeepSeek\_V3 685B        & 0.245     & 1       & ns \\
DeepSeek\_R1 685B     & DeepSeek\_V3 685B        & 1         & 1       & ns \\
\hline
\end{tabular}%
}
\end{table}

\begin{table}[ht]
\centering
\caption{Post hoc test results for Combinatorics under the Spanish\_ZS configuration. Adjusted p-values and significance levels are reported.}\label{Tab:PosthoctestresultsforCombonatoricsundertheSpanish_ZSconfiguration.}
\resizebox{\textwidth}{!}{%
\begin{tabular}{|l|l|c|c|c|}
\hline
\textbf{Group 1} & \textbf{Group 2} & \textbf{p} & \textbf{p.adj} & \textbf{p.adj.signif} \\
\hline
GPT\_4o          & GPT\_4o\_mini       & 0.0673    & 1       & ns \\
GPT\_4o          & o3\_mini            & 1         & 1       & ns \\
GPT\_4o\_mini     & o3\_mini            & 0.0673    & 1       & ns \\
GPT\_4o          & LLaMA\_3\_3\_70B     & 0.141     & 1       & ns \\
GPT\_4o\_mini     & LLaMA\_3\_3\_70B     & 1         & 1       & ns \\
o3\_mini         & LLaMA\_3\_3\_70B     & 0.141     & 1       & ns \\
GPT\_4o          & DeepSeek\_R1 685       & 1         & 1       & ns \\
GPT\_4o\_mini     & DeepSeek\_R1 685B       & 0.0673    & 1       & ns \\
o3\_mini         & DeepSeek\_R1 685B       & 1         & 1       & ns \\
LLaMA\_3\_3\_70B  & DeepSeek\_R1 685B        & 0.141     & 1       & ns \\
GPT\_4o          & DeepSeek\_V3 685B        & 1         & 1       & ns \\
GPT\_4o\_mini     & DeepSeek\_V3 685B        & 0.0673    & 1       & ns \\
o3\_mini         & DeepSeek\_V3 685B        & 1         & 1       & ns \\
LLaMA\_3\_3\_70B  & DeepSeek\_V3 685B        & 0.141     & 1       & ns \\
DeepSeek\_R1 685B     & DeepSeek\_V3 685B        & 1         & 1       & ns \\
\hline
\end{tabular}%
}
\end{table}

\begin{table}[ht]
\centering
\caption{Post hoc test results for Probability under the Spanish\_ZS\_CoT configuration. Adjusted p-values and significance levels are reported.}\label{Tab:PosthoctestresultsforProbabilityundertheSpanish_ZS_CoTconfiguration.}
\resizebox{\textwidth}{!}{%
\begin{tabular}{|l|l|c|c|c|}
\hline
\textbf{Group 1} & \textbf{Group 2} & \textbf{p} & \textbf{p.adj} & \textbf{p.adj.signif} \\
\hline
GPT\_4o          & GPT\_4o\_mini       & 6.83E-01  & 1         & ns   \\
GPT\_4o          & o3\_mini            & 1.46E-03  & 0.0176    & *    \\
GPT\_4o\_mini     & o3\_mini            & 1.26E-02  & 0.101     & ns   \\
GPT\_4o          & LLaMA               & 1.00E+00  & 1         & ns   \\
GPT\_4o\_mini     & LLaMA               & 1.00E+00  & 1         & ns   \\
o3\_mini         & LLaMA               & 4.60E-03  & 0.046     & *    \\
GPT\_4o          & DeepSeek\_R1 685B       & 1.46E-03  & 0.0176    & *    \\
GPT\_4o\_mini     & DeepSeek\_R1 685B        & 1.26E-02  & 0.101     & ns   \\
o3\_mini         & DeepSeek\_R1 685B        & 1.00E+00  & 1         & ns   \\
LLaMA           & DeepSeek\_R1 685B        & 4.60E-03  & 0.046     & *    \\
GPT\_4o          & DeepSeek\_V3 685B        & 8.86E-05  & 0.00133   & **   \\
GPT\_4o\_mini     & DeepSeek\_V3 685B        & 1.02E-03  & 0.0132    & *    \\
o3\_mini         & DeepSeek\_V3 685B        & 5.93E-01  & 1         & ns   \\
LLaMA           & DeepSeek\_V3 685B        & 3.19E-04  & 0.00447   & **   \\
DeepSeek\_R1 685B     & DeepSeek\_V3 685B        & 5.93E-01  & 1         & ns   \\
\hline
\end{tabular}%
}
\end{table}

\begin{table}[ht]
\centering
\caption{Post hoc test results for Geometry under the Spanish\_ZS\_CoT configuration. Adjusted p-values and significance levels are reported.}\label{Tab:PosthoctestresultsforGeometryundertheSpanish_ZS_CoTconfiguration.}
\resizebox{\textwidth}{!}{%
\begin{tabular}{|l|l|c|c|c|}
\hline
\textbf{Group 1} & \textbf{Group 2} & \textbf{p} & \textbf{p.adj} & \textbf{p.adj.signif} \\
\hline
GPT\_4o          & GPT\_4o\_mini       & 0.709     & 1       & ns \\
GPT\_4o          & o3\_mini            & 0.0528    & 0.686   & ns \\
GPT\_4o\_mini     & o3\_mini            & 0.00909   & 0.136   & ns \\
GPT\_4o          & LLaMA               & 1         & 1       & ns \\
GPT\_4o\_mini     & LLaMA               & 0.461     & 1       & ns \\
o3\_mini         & LLaMA               & 0.111     & 1       & ns \\
GPT\_4o          & DeepSeek\_R1 685B        & 0.264     & 1       & ns \\
GPT\_4o\_mini     & DeepSeek\_R1 685B        & 0.0673    & 0.807   & ns \\
o3\_mini         & DeepSeek\_R1 685B        & 0.648     & 1       & ns \\
LLaMA           & DeepSeek\_R1 685B        & 0.449     & 1       & ns \\
GPT\_4o          & DeepSeek\_V3 685B        & 0.13      & 1       & ns \\
GPT\_4o\_mini     & DeepSeek\_V3 685B        & 0.0271    & 0.379   & ns \\
o3\_mini         & DeepSeek\_V3 685B        & 1         & 1       & ns \\
LLaMA           & DeepSeek\_V3 685B        & 0.245     & 1       & ns \\
DeepSeek\_R1 685B     & DeepSeek\_V3 685B        & 1         & 1       & ns \\
\hline
\end{tabular}%
}
\end{table}

\begin{table}[ht]
\centering
\caption{Post hoc test results for Calculus under the Spanish\_ZS\_CoT configuration. Adjusted p-values and significance levels are reported.}\label{Tab:PosthoctestresultsforCalculusundertheSpanish_ZS_CoTconfiguration.}
\resizebox{\textwidth}{!}{%
\begin{tabular}{|l|l|c|c|c|}
\hline
\textbf{Group 1} & \textbf{Group 2} & \textbf{p} & \textbf{p.adj} & \textbf{p.adj.signif} \\
\hline
GPT\_4o          & GPT\_4o\_mini       & 1         & 1       & ns \\
GPT\_4o          & o3\_mini            & 0.023     & 0.345   & ns \\
GPT\_4o\_mini     & o3\_mini            & 0.023     & 0.345   & ns \\
GPT\_4o          & LLaMA               & 0.648     & 1       & ns \\
GPT\_4o\_mini     & LLaMA               & 0.648     & 1       & ns \\
o3\_mini         & LLaMA               & 0.141     & 1       & ns \\
GPT\_4o          & DeepSeek\_R1 685B        & 0.388     & 1       & ns \\
GPT\_4o\_mini     & DeepSeek\_R1 685B        & 0.388     & 1       & ns \\
o3\_mini         & DeepSeek\_R1 685B        & 0.272     & 1       & ns \\
LLaMA           & DeepSeek\_R1 685B        & 1         & 1       & ns \\
GPT\_4o          & DeepSeek\_V3 685B        & 0.0528    & 0.686   & ns \\
GPT\_4o\_mini     & DeepSeek\_V3 685B        & 0.0528    & 0.686   & ns \\
o3\_mini         & DeepSeek\_V3 685B        & 1         & 1       & ns \\
LLaMA           & DeepSeek\_V3 685B        & 0.264     & 1       & ns \\
DeepSeek\_R1 685B     & DeepSeek\_V3 685B        & 0.461     & 1       & ns \\
\hline
\end{tabular}%
}
\end{table}

\end{document}